%% file: main.tex
\pdfoutput=1
\documentclass[11pt]{article}

\PassOptionsToPackage{table}{xcolor}
\usepackage[preprint]{acl}
\usepackage{times}
\usepackage{latexsym}
\usepackage[T1]{fontenc}
\usepackage[utf8]{inputenc}
\usepackage{microtype}
\usepackage{inconsolata}
\usepackage{graphicx}
\usepackage{booktabs}
\usepackage{tabularx}
\usepackage{amsmath,amssymb}
\usepackage{multirow}
\usepackage{balance}
\usepackage{etoolbox}
\apptocmd{\thebibliography}{\setlength{\itemsep}{1pt plus 3pt minus 0.5pt}\setlength{\parsep}{1pt plus 2pt}}{}{}

\definecolor{tblhi}{HTML}{E6F0FA}
\definecolor{tblpos}{HTML}{EAF6EC}
\definecolor{tblposhi}{HTML}{CFEAD5}
\definecolor{tblneg}{HTML}{FDECEB}
\definecolor{tblzebra}{HTML}{F7F8FA}
\definecolor{tblsub}{HTML}{F0F3F8}
\definecolor{posfg}{HTML}{1B6E35}
\definecolor{negfg}{HTML}{B3261E}
\definecolor{bestfg}{HTML}{0B4F8C}
\newcommand{\hirow}{\rowcolor{tblhi}}
\newcommand{\zrow}{\rowcolor{tblzebra}}

\newcommand{\cbest}{\cellcolor{tblhi}}
\newcommand{\cpos}{\cellcolor{tblpos}}
\newcommand{\cposhi}{\cellcolor{tblposhi}}
\newcommand{\cneg}{\cellcolor{tblneg}}
\newcommand{\tpos}[1]{\textcolor{posfg}{#1}}
\newcommand{\tneg}[1]{\textcolor{negfg}{#1}}
\newcommand{\tbest}[1]{\textcolor{bestfg}{#1}}
\newcommand{\cmark}{\textcolor{posfg}{$\checkmark$}}
\newcommand{\xmark}{\textcolor{negfg}{$\times$}}
\newcommand{\fstar}{\textcolor{bestfg}{$\star$}}

\newcolumntype{Y}{>{\raggedright\arraybackslash}X}

\title{Reliability-Aware Multimodal Sentiment Prediction Under Local Modality Missingness}

\author{
  \textbf{Minghao Kong\textsuperscript{1,2}, Yao Deng\textsuperscript{3},
  Xi Zhang\textsuperscript{4}, Ying Gao\textsuperscript{2}} \\
  \textbf{Jiurun Chen\textsuperscript{2}, Zirong Liu\textsuperscript{2},
  Jianxin Wu\textsuperscript{1}\textsuperscript{\hyperlink{corrauth}{\ensuremath{*}}},
  Rongjie Wang\textsuperscript{2}\textsuperscript{\hyperlink{corrauth}{\ensuremath{*}}}} \\[0.5em]
  {\large\normalfont\textsuperscript{1}School of Electronics and Communication Engineering,} \\
  {\large\normalfont Sun Yat-sen University, Shenzhen, China} \\
  {\large\normalfont\textsuperscript{2}Department of Network Intelligence, Pengcheng Laboratory, Shenzhen, China} \\
  {\large\normalfont\textsuperscript{3}School of Mathematics, Sun Yat-sen University, Guangzhou, China} \\
  {\large\normalfont\textsuperscript{4}School of Physics and Astronomy, Sun Yat-sen University, Zhuhai, China}
}
\hypersetup{pdfauthor={Minghao Kong, Yao Deng, Xi Zhang, Ying Gao, Jiurun Chen, Zirong Liu, Jianxin Wu, Rongjie Wang}}

\begin{document}
\maketitle
{\renewcommand{\thefootnote}{\fnsymbol{footnote}}%
\footnotetext[1]{\raisebox{\baselineskip}[0pt][0pt]{\hypertarget{corrauth}{}}Corresponding authors.}}

\begin{abstract}
Multimodal sentiment models are typically evaluated on complete utterances, yet deployed streams often lose contiguous spans of text, audio, or video. We study such local missingness on aligned CMU-MOSEI features. Our reliability-aware network (RM-Net) preserves original temporal positions, marks unobserved positions with a learned token, outputs bounded per-modality intensity contributions gated by observation quality, and trains with dual-view contiguous-span masking. At inference, we fuse five-seed ensembles of RM-Net and a LoRA-adapted BERT text expert using deterministic weights governed solely by the visible fraction of content tokens. All architecture, checkpoint, and fusion choices are selected and frozen on validation. On the held-out split, the fused predictor achieves a Macro-F1 of 0.655 on clean inputs and 0.634 under 40\% mixed missingness; its two-view average exceeds the RM-Net auxiliary-head baseline by 0.024 (95\% video-cluster bootstrap CI [0.003, 0.045]). Because this split was inspected during development, these numbers are not a blind test. Exact Shapley attributions across all eight modality subsets and position-level occlusion on the deployed predictor show that removing top-ranked positions lowers predicted-class probability faster than random removal (AUC gap 0.113 on validation). We also document an independent, auditable word-level alignment pipeline for raw clips whose outputs do not feed the predictive models.
\end{abstract}

\section{Introduction}

Speakers convey sentiment jointly through lexical content, prosody, and facial behavior, and multimodal sentiment models exploit these complementary cues \citep{zadeh2017tfn,tsai2019mult}. Most models assume every modality remains available throughout an utterance. In practice, this assumption fails in a structured way: a microphone drops out for several seconds, speech recognition misses a phrase, or a speaker's face leaves the frame. Unlike the whole-modality loss or random corruption emphasized in much prior work, such \emph{local, contiguous} missingness leaves surviving modalities informative and requires preserving their temporal alignment across the gap.

Handling local missingness raises three linked requirements. First, absent spans must be modeled explicitly rather than conflated with observed content: without an observation mask, z-scored zero-filling treats an absent face as a mean-valued observation. Second, fusion must reflect available evidence so that a text expert's influence contracts as words disappear. Third, auditing degraded predictions requires intervention-based explanations directly on the deployed predictor, since raw attention weights are not faithful explanations.

We address these requirements with a modular framework combining a position-preserving multimodal core (RM-Net), an availability-gated LoRA BERT text expert, and exact intervention-based explanations. Because predictive benchmarks use pre-aligned token features, we also document a separate word-level alignment pipeline for raw clips. Our main contributions are:
\begin{itemize}
\item \textbf{Observation-aware multimodal core:} RM-Net preserves temporal indices, marks gaps with learned tokens, gates modalities by observation quality, and trains with dual-view contiguous masking;
\item \textbf{Availability-gated fusion:} deterministic visible-text gating combines RM-Net with a LoRA-adapted BERT expert, with all settings frozen on validation;
\item \textbf{Controlled robustness evaluation:} evaluation across 21 missingness conditions, span position/length diagnostics, ablations, and a pre-encoding text-span deletion check, disclosing earlier held-out inspection;
\item \textbf{Dual-target explanations:} exact eight-subset Shapley attributions and position-level occlusion for intensity and class probability, audited with deletion/retention curves and class flips;
\item \textbf{Auditable raw-clip alignment:} an independent, reproducible word-level alignment pipeline for raw clips, kept separate from the predictive models.
\end{itemize}

\section{Related Work}

\paragraph{Multimodal sentiment fusion.}
Foundational sentiment architectures combine lexical, acoustic, and visual cues via tensor products \citep{zadeh2017tfn}, low-rank modality factors \citep{liu2018lmf}, or modality-invariant and -specific subspaces \citep{hazarika2020misa}, while cross-modal Transformers model asynchronous streams directly \citep{tsai2019mult}. These models generally assume complete inputs and are benchmarked on CMU-MOSEI \citep{bagher-zadeh-etal-2018-multimodal}.

\paragraph{Incomplete multimodal data.}
Recent methods handle incomplete sentiment inputs through reconstruction, proxy, or noise-robust mechanisms \citep{zhang2024lnln,zhu2025prmf,sun2024emtdlfr}, including random within-modality feature missingness \citep{zhu2025prmf,sun2024emtdlfr}. We differ in structure and modularity: we examine \emph{contiguous} spans of varying length and placement while preserving original temporal indices, keeping pretrained encoders frozen and fusing a separately trained text expert at inference.

\paragraph{Parameter-efficient adaptation.}
Low-rank adaptation (LoRA; \citealp{hu2022lora}) inserts compact trainable updates into a frozen Transformer. Applying LoRA to BERT \citep{devlin-etal-2019-bert} yields a task-adapted text expert trained solely on the training split while keeping the multimodal core decoupled.

\paragraph{Explanations and uncertainty.}
Shapley values \citep{shapley1953value} and SHAP \citep{lundberg2017shap} attribute predictions to feature groups, and deletion/insertion curves test whether top-ranked evidence drives outputs \citep{petsiuk2018rise}. Because attention weights are unreliable explanations \citep{jain2019attention} and gradient saliency can fail sanity checks \citep{adebayo2018sanity}, we intervene directly on the deployed predictor. Although we ensemble five seeds, ensemble uncertainty estimation \citep{lakshminarayanan2017ensembles} and calibration \citep{guo2017calibration} lie outside our scope; our probabilities are uncalibrated.

\section{Data and Input Representation}
\label{sec:data}

\paragraph{Dataset, splits, and labels.}
Predictive experiments use a supplied 4,850-utterance subset derived from CMU-MOSEI \citep{bagher-zadeh-etal-2018-multimodal} with its provided partition (the full corpus has roughly 23,500 labeled utterances). The training, validation, and held-out splits contain 3,395, 728, and 727 utterances, with validation spanning 239 source videos. Continuous labels $y\in[-3,3]$ map to Negative ($y<0$), Neutral ($y=0$; 22.3\% of training), and Positive ($y>0$). Two unlabeled sets of 30 and 20 utterances serve strictly for inference and descriptive explanation analysis.

\paragraph{Aligned multimodal features.}
Each utterance spans 50 aligned positions per modality. Text positions correspond to BERT WordPiece tokens (including \texttt{[CLS]} and \texttt{[SEP]}), not surface words. Acoustic and visual inputs are supplied 74-dimensional and 35-dimensional pre-extracted vectors; we neither produced them nor verified their extraction toolchain. We re-encode text with a frozen uncased BERT-base model \citep{devlin-etal-2019-bert} at a fixed revision so both branches share identical text inputs; on observed positions, re-encoded vectors match supplied text features within a mean absolute difference of roughly $3\times10^{-4}$ (consistent with half-precision storage).

\paragraph{Observation masks.}
Let $\ell\le 50$ be the valid position count of a modality, with $p_i=\mathbf 1\{0\le i<\ell\}$. A position is observed when its raw vector is non-zero: $o_i^{m}=p_i\mathbf 1\{\|x_i^m\|_0>0\}$. Testing raw vectors before normalization separates missing spans from z-scored values near zero, though legitimate all-zero states (e.g., natural silence or absent faces) are also treated as unobserved. Features are z-scored using training statistics, clipped to $[-5,5]$, and uninformative positions are reset to zero. A synthetic missingness set $\mathcal M_m$ (a union of half-open spans) yields the damaged mask $o_i^{m,c}=o_i^{m}\mathbf 1\{i\notin\mathcal M_m\}$; positions are never dropped, preserving cross-modal alignment.

\section{Method}
\label{sec:method}

\begin{figure*}[t]
\centering
\includegraphics[width=\textwidth]{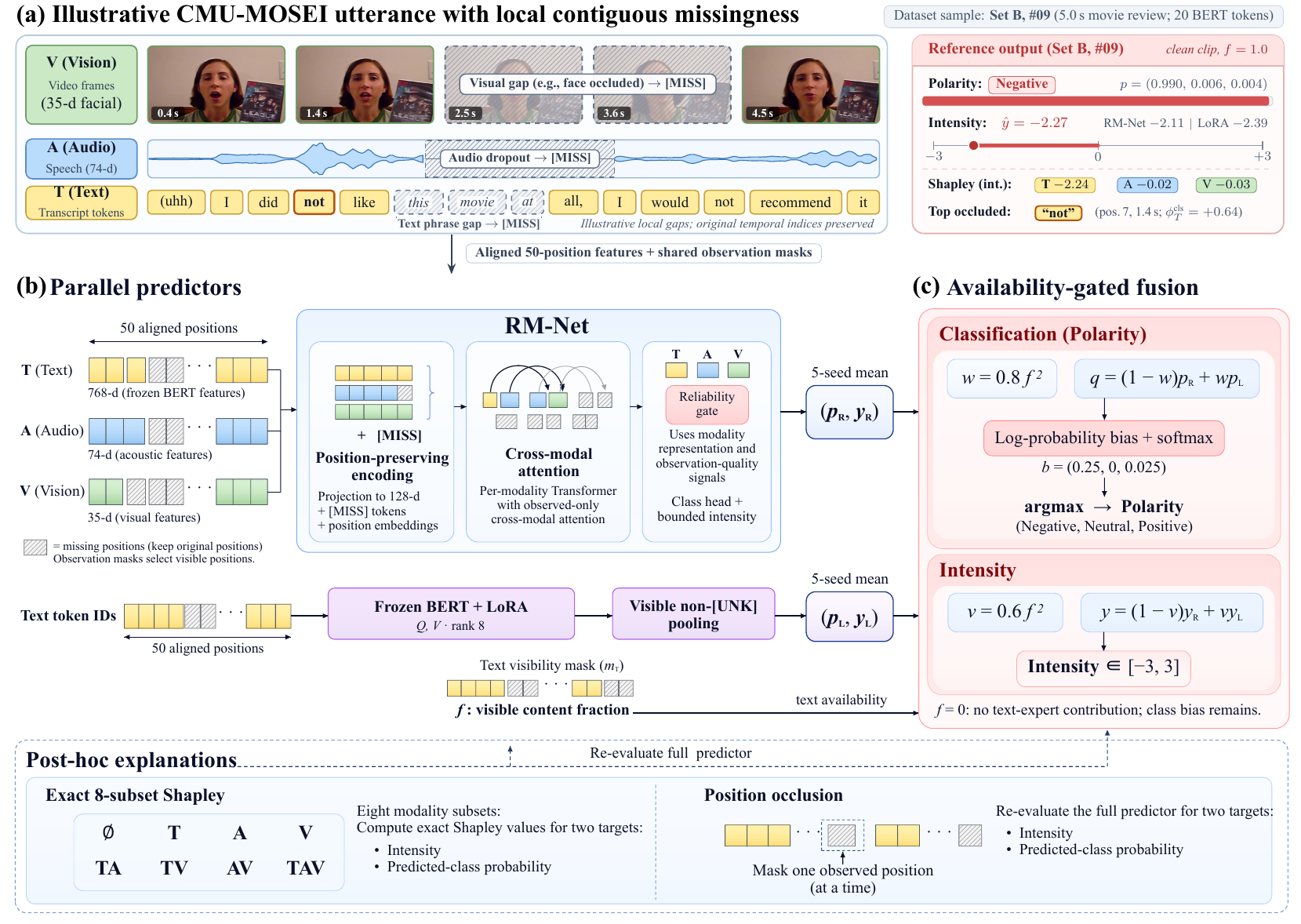}
\caption{Overview of the fused predictor. (a)~An illustrative CMU-MOSEI utterance (Set~B, \#09) showing real video frames, speech waveform, and transcript tokens with illustrative contiguous missing spans (left) alongside the model's unperturbed reference output and attributions on the clean clip (right). (b)~Parallel RM-Net and LoRA text-expert branches evaluated under shared observation masks. (c)~Availability-gated fusion driven solely by visible text fraction $f$, audited via exact Shapley and position-occlusion interventions on the full predictor.}
\label{fig:overview}
\end{figure*}

Figure~\ref{fig:overview} summarizes our pipeline. RM-Net and the LoRA text expert are trained separately (five seeds each) and fused only at inference under shared masks (full equations in Appendix~\ref{app:model}).

\subsection{RM-Net core}
Each modality is projected to 128 dimensions, unobserved positions are replaced by a learned \texttt{[MISS]} embedding, and learned positional embeddings are added. A two-layer, four-head Transformer \citep{vaswani2017attention} (FFN width 256) encodes each modality while masking only padding, keeping internal gaps intact. A cross-modal layer then lets each position attend solely to observed positions of the other modalities with a learned negative distance bias $b_m(i,j)=-\operatorname{softplus}(\beta_m)|i-j|/5$. Attention pooling over observed positions yields a modality vector $z_m$, falling back to a learned absent vector when no position is observed.

A reliability gate combines $z_m$ with two observation descriptors---unobserved fraction $\rho_m$ and normalized unobserved-segment count---and a stop-gradient reconstruction residual $\epsilon_m$ computed strictly on observed positions. Normalized weights $w_m$ over available modalities produce an additive, bounded intensity:
\begin{equation*}
\begin{aligned}
\hat y_m&=2.7\tanh\bigl(a_m(z_m)\bigr),\\
\hat y_{\mathrm{int}}&=0.3\tanh\bigl(a_{\mathrm{int}}(z)\bigr),\\
\hat y&=\sum_m w_m\hat y_m+\hat y_{\mathrm{int}},
\end{aligned}
\end{equation*}
where $z=\sum_m w_m z_m$. When no modality is observed, a learned bounded prior replaces $\hat y_{\mathrm{int}}$, guaranteeing $\hat y\in[-3,3]$. Each term $w_m\hat y_m$ is exact, giving an additive decomposition by construction. An auxiliary head maps $z$ to three class logits. The core has 1,510,206 parameters ($\approx$3 MB in FP16).

\subsection{Dual-view training with contiguous masking}
Each batch runs twice under shared parameters: clean and masked for augmented samples. The objective is
\begin{equation*}
\begin{aligned}
\mathcal L&=\mathcal L_{\mathrm{pred}}+0.3\mathcal L_{u}+0.3\mathcal L_{\mathrm{CE}}\\
&\quad+0.5\mathcal L_{\mathrm{rec}}+0.5\mathcal L_{\mathrm{con}}+0.1\mathcal L_{\mathrm{int}}.
\end{aligned}
\end{equation*}
Supervised terms include L1 intensity loss on both views ($\mathcal L_{\mathrm{pred}}$), single-modality L1 loss on surviving modalities in the damaged view ($\mathcal L_{u}$), and class-weighted cross-entropy on both views ($\mathcal L_{\mathrm{CE}}$). Regularizers reconstruct newly masked positions that were observed in the clean view ($\mathcal L_{\mathrm{rec}}$), pull damaged fused representations and intensities toward stop-gradient clean outputs ($\mathcal L_{\mathrm{con}}$), and penalize damaged-view interaction magnitude ($\mathcal L_{\mathrm{int}}$).

In each batch, 20\% of samples stay unaugmented; otherwise 1, 2, or 3 modalities are masked with probabilities 0.5, 0.35, and 0.15. Missing ratios follow $0.7\,U(0.1,0.6)+0.25\,U(0.6,0.95)+0.05\,\delta(1)$ in one or two contiguous spans, with the ratio cap ramping linearly from 0.3 to 1 over the first 30\% of training. Separately, 1\% of augmented samples mask all modalities to supervise the empty-set prior.

\subsection{Text expert and availability-gated fusion}
To supply task-adapted lexical cues, the text expert freezes BERT-base and adds rank-8 LoRA updates ($\alpha=16$) to query and value projections in all 12 layers ($\approx$0.29M trainable adapter parameters, excluding heads). Mean pooling over visible non-\texttt{[UNK]} positions feeds a shared 128-unit hidden layer with separate classification and intensity heads, trained with clean/masked class-weighted cross-entropy plus $0.3$ times L1 intensity loss.

Let $f$ be the visible fraction of content positions (excluding \texttt{[PAD]}, \texttt{[CLS]}, and \texttt{[SEP]}; the numerator also excludes \texttt{[UNK]}, with $f=0$ if none exist). With five-seed mean class probabilities and intensities $(p_R, y_R)$ and $(p_L, y_L)$, fusion applies
\begin{align*}
w&=0.8f^2, & q&=(1-w)p_R+wp_L,\\
v&=0.6f^2, & p_c&\propto e^{b_c}\max(q_c,10^{-12}),\\
\hat c&=\operatorname*{arg\,max}_c p_c, & \hat y&=(1-v)y_R+vy_L,
\end{align*}
with bias $b=(0.25,\,0,\,0.025)$ for (Negative, Neutral, Positive). When $f=0$, expert weights vanish: intensity reverts to RM-Net while classification keeps $b$. Polarity and intensity are fused separately and may disagree; $b$ adjusts decisions rather than calibrating probabilities.

\subsection{Explanations on the fused predictor}
We audit the deployed predictor at modality and position levels. Treating text, audio, and vision as three Shapley players, we keep only observations in each of the eight subsets $S$ and re-run the entire fused predictor---including expert pooling and availability weights---to compute exact Shapley values for fused intensity $\hat y$ and the fixed originally predicted class probability. For position-level evidence, each observed content position is occluded individually and ranked by $|\Delta y|$ or predicted-class probability drop. We evaluate deletion/retention curves at fractions $0,0.1,\dots,0.8$ against five random permutations and record the first class-flip deletion step.

\begin{table*}[t]
\centering
\small
\renewcommand{\arraystretch}{1.08}
\setlength{\tabcolsep}{7pt}
\begin{tabular}{lccccc}
\toprule
& \multicolumn{3}{c}{\textbf{Validation Macro-F1 ($\uparrow$)}} & \multicolumn{2}{c}{\textbf{Held-out Macro-F1 ($\uparrow$)}}\\
\cmidrule(lr){2-4}\cmidrule(lr){5-6}
\textbf{Configuration} & Clean & Mixed40 & Avg & Clean & Mixed40\\
\midrule
RM-Net auxiliary head & 0.6144 & 0.6057 & 0.6101 & 0.6237 & 0.6178\\
\zrow
RM-Net, CE weight 1, 15 seeds, bias & 0.6260 & 0.6277 & 0.6269 & 0.6187 & 0.6071\\
RM-Net + frozen BERT probe (15 seeds), bias & \textbf{0.6360} & \underline{0.6367} & \underline{0.6363} & 0.6326 & 0.6284\\
\midrule
\zrow
LoRA text expert only & 0.6175 & 0.6269 & 0.6222 & \underline{0.6578} & \textbf{0.6524}\\
RM-Net + LoRA, no bias & \underline{0.6351} & 0.6304 & 0.6328 & \textbf{0.6600} & \underline{0.6496}\\
\hirow
\fstar\;\textbf{RM-Net + LoRA, with bias (final)} & 0.6339 & \textbf{0.6394} & \textbf{0.6367} & 0.6550 & 0.6341\\
\bottomrule
\end{tabular}
\caption{Macro-F1 of key development configurations. Best scores per column are \textbf{bold}, second-best are \underline{underlined}, and \fstar{} marks the final model selected strictly by validation average. Held-out columns were inspected during development and do not represent a blind test. The frozen-probe row pairs a 15-seed core (CE weight 1) with frozen BERT features, whereas the final model uses five core seeds.}
\label{tab:comparison}
\end{table*}

\section{Experimental Protocol}
\label{sec:protocol}

\paragraph{Selection and frozen decisions.}
All architecture, checkpoint, fusion, and bias settings were chosen on validation. RM-Net checkpoints were selected via $J=r-\mathrm{MAE}+\mathrm{Macro\text{-}F}_1$ ($r$: Pearson correlation), weighting the clean view and the mean across 20\%, 40\%, and 60\% missingness at 0.5 each, with Macro-F1 intensity cut-points re-tuned on validation predictions per epoch. The LoRA expert and fusion rule maximized the validation mean of clean and 40\% mixed-missingness Macro-F1, searching weights on $[0,1]$ (step 0.1), exponents on $\{1,2,4\}$, and class biases on $[-1,1]$ (step 0.1, refined to 0.025). Settings were frozen before final held-out evaluation; however, earlier held-out inspection means these results are not a blind estimate (candidates in Appendix~\ref{app:selection}).

\paragraph{Missingness conditions.}
``Clean'' inputs retain only natural zero vectors. Under ``mixed40'', each utterance receives a damaged subset from the seven non-empty combinations (T, A, V, TA, TV, AV, TAV), losing one contiguous span of $\lceil0.4\ell\rceil$ positions per damaged modality (seed 20260923); thus 40\% is a per-modality rate, not the share of fully missing utterances. The 21-condition grid crosses all seven subsets with 20\%, 40\%, and 60\% rates. Because feature masking occurs after BERT encoding, surviving tokens may retain contextual leakage from masked words; Section~\ref{sec:pre-encoding} tests this directly.

\paragraph{Metrics and statistics.}
We report Accuracy, Macro-F1, MAE, and Pearson $r$ using $\operatorname*{arg\,max}_c p_c$ and $\hat y$. Because utterances from the same video are correlated, primary comparisons use 5,000 paired video-cluster bootstrap resamples \citep{efron1979bootstrap} (retaining both views per sampled video) alongside a two-sided video-cluster permutation test. By contrast, Table~\ref{tab:fidelity} intervals use 500 utterance-level paired bootstrap resamples (keeping each curve intact) without video clustering.

\paragraph{Compute.}
RM-Net checkpoints were trained on four RTX 3090 GPUs in BF16; frozen BERT-base has roughly 110M parameters. Cumulative development GPU hours were not logged and are not reported.

\section{Results}
\label{sec:results}

\subsection{Model comparison on validation and held-out splits}
Table~\ref{tab:comparison} separates core, text-representation, and fusion choices. The final configuration achieves the top validation average across all 29 candidates (Appendix~\ref{app:candidates}), edging out the strongest frozen-probe fusion by 0.0003 on unrounded validation scores. Because that comparator uses a 15-seed core (CE weight 1) rather than our five-seed core, this does not show general superiority of LoRA over frozen features. Several candidates lead on individual held-out views---such as LoRA-only on mixed40 and unbiased fusion on clean inputs---and class biases aid validation without improving held-out ranking, marking their gain as selection-dependent.

\begin{table}[t]
\centering
\small
\renewcommand{\arraystretch}{1.08}
\setlength{\tabcolsep}{4.5pt}
\begin{tabular}{lcccc}
\toprule
& \multicolumn{2}{c}{\textbf{Clean}} & \multicolumn{2}{c}{\textbf{Mixed40}}\\
\cmidrule(lr){2-3}\cmidrule(lr){4-5}
\textbf{Metric} & Core & Fused & Core & Fused\\
\midrule
Acc & 0.6479 & \cbest\textbf{0.6809} & 0.6424 & \cbest\textbf{0.6630}\\
\zrow
Macro-F1 & 0.6237 & \cbest\textbf{0.6550} & 0.6178 & \cbest\textbf{0.6341}\\
MAE & 0.6344 & \cbest\textbf{0.6099} & 0.6379 & \cbest\textbf{0.6210}\\
\zrow
$r$ & 0.6644 & \cbest\textbf{0.6942} & 0.6677 & \cbest\textbf{0.6861}\\
\bottomrule
\end{tabular}
\caption{Held-out performance (727 utterances). Higher is better ($\uparrow$) for Acc, Macro-F1, and $r$; lower is better ($\downarrow$) for MAE. Shaded \textbf{bold} cells mark the better model in each condition. Core is RM-Net with its auxiliary classification head and intensity output; Fused is RM-Net + LoRA.}
\label{tab:final}
\end{table}

\subsection{Final model on the held-out split}
Table~\ref{tab:final} compares the fused model against RM-Net core, showing gains on all four metrics across both views. Macro-F1 rises from 0.6237 to 0.6550 on clean inputs and from 0.6178 to 0.6341 under mixed40. The two-view average increases by 0.0238, from 0.6208 to 0.6446 (95\% CI $[0.0027,0.0452]$; two-sided permutation $p=0.030$). Individual clean and mixed40 CIs are $[0.0049,0.0585]$ and $[-0.0043,0.0371]$, supporting the two-view average more clearly than mixed40 alone.

\begin{table}[t]
\centering
\small
\renewcommand{\arraystretch}{1.08}
\setlength{\tabcolsep}{2.5pt}
\begin{tabular}{lccc@{\hspace{7pt}}ccc}
\toprule
& \multicolumn{3}{c}{\textbf{$\Delta$ Macro-F1 (pts, $\uparrow$)}} & \multicolumn{3}{c}{\textbf{$\Delta$ MAE red.\ ($\uparrow$)}}\\
\cmidrule(lr){2-4}\cmidrule(lr){5-7}
\textbf{Sub.} & 20\% & 40\% & 60\% & 20\% & 40\% & 60\%\\
\midrule
T & \cpos\tpos{+0.74} & \cpos\tpos{+0.69} & \cneg\textbf{\tneg{$-$0.30}} & \cpos\tpos{0.017} & \cpos\tpos{0.012} & \cpos\tpos{0.006}\\
A & \cposhi\textbf{\tpos{+3.15}} & \cposhi\textbf{\tpos{+3.33}} & \cposhi\textbf{\tpos{+2.87}} & \cposhi\textbf{\tpos{0.024}} & \cposhi\textbf{\tpos{0.024}} & \cposhi\textbf{\tpos{0.023}}\\
V & \cposhi\textbf{\tpos{+3.32}} & \cposhi\textbf{\tpos{+3.41}} & \cposhi\textbf{\tpos{+3.06}} & \cposhi\textbf{\tpos{0.024}} & \cposhi\textbf{\tpos{0.022}} & \cposhi\textbf{\tpos{0.022}}\\
TA & \cpos\tpos{+1.73} & \cpos\tpos{+1.31} & \cpos\tpos{+0.70} & \cpos\tpos{0.019} & \cpos\tpos{0.012} & \cpos\tpos{0.005}\\
TV & \cpos\tpos{+2.32} & \cpos\tpos{+1.85} & \cpos\tpos{+0.21} & \cpos\tpos{0.018} & \cpos\tpos{0.012} & \cpos\tpos{0.005}\\
AV & \cposhi\textbf{\tpos{+3.33}} & \cposhi\textbf{\tpos{+3.40}} & \cposhi\textbf{\tpos{+3.66}} & \cposhi\textbf{\tpos{0.023}} & \cposhi\textbf{\tpos{0.022}} & \cposhi\textbf{\tpos{0.020}}\\
TAV & \cpos\tpos{+2.10} & \cpos\tpos{+1.03} & \cneg\textbf{\tneg{$-$1.44}} & \cposhi\textbf{\tpos{0.020}} & \cpos\tpos{0.010} & \cpos\tpos{0.005}\\
\bottomrule
\end{tabular}
\caption{Changes from the RM-Net auxiliary head to the fused model on the 727-utterance held-out split across 21 fixed missingness conditions. Green shading marks gains (darker \textbf{bold}: $\ge$2.5 pts or $\ge$0.020 MAE reduction); red marks declines. Cells are single conditions, not independent samples.}
\label{tab:grid}
\end{table}

\subsection{Missingness grid}
Table~\ref{tab:grid} reports held-out gains across all 21 missingness conditions. Fusion improves Macro-F1 in 19 conditions and reduces MAE in all 21. Gains peak when text stays intact (+2.87 to +3.66 Macro-F1 points for A, V, and AV, alongside the largest MAE reductions) and narrow for TA and TV. Both Macro-F1 drops occur at 60\% text masking: T ($-$0.30 points) and TAV ($-$1.44 points).

Validation diagnostics mirror this lexical sensitivity. Masking text at the head, middle, or tail increases MAE by 0.0214, 0.0165, and 0.0287, versus at most 0.0022 for audio or vision (descriptive across distinct utterance groups). Text spans of $\ge$13 positions cause the largest MAE increase (0.0343), though missing rate and utterance length confound absolute span length. Under total text loss, held-out Macro-F1 drops to 0.454; with all modalities absent, output reverts to the learned prior (Macro-F1 0.148). Appendix~\ref{app:core} reports core scans, and Appendix~\ref{app:fused} details position, length, and replay analyses.

\begin{table}[t]
\centering
\small
\renewcommand{\arraystretch}{1.08}
\begin{tabularx}{\linewidth}{@{}Yc@{}}
\toprule
\textbf{Variant} & \textbf{Val MAE ($\downarrow$)}\\
\midrule
\hirow
\fstar\;\textbf{RM-Net (full)} & \textbf{0.5970}\,$\pm$\,\textbf{0.0046}\\
\midrule
$-$ Masking augmentation (clean only) & \cneg\tneg{$0.6065\pm0.0071$}\\
\zrow
$-$ Consistency loss & $0.6023\pm0.0059$\\
$-$ Reliability gate (concat.\ MLP) & $0.6026\pm0.0092$\\
\zrow
$-$ Audio and vision (text only) & $0.6014\pm0.0068$\\
\bottomrule
\end{tabularx}
\caption{Core-network ablations on the validation missingness grid (mean $\pm$ SD across five seeds). \fstar{} highlights the full model; red shading marks the largest degradation.}
\label{tab:ablation}
\end{table}

\subsection{Ablations of the core network}
Table~\ref{tab:ablation} reports validation MAE over the missingness grid for core variants, averaging each seed over subsets, rates, and placements before summarizing across the five seeds. Removing masking augmentation worsens MAE the most (+0.0095); removing consistency loss or replacing gated fusion with a concatenation MLP worsens MAE by $\approx$0.005 each, and text-only input worsens MAE by 0.0044. Because these shifts are comparable to seed-to-seed SDs, we do not claim significant individual component effects.

\begin{figure*}[t]
\centering
\includegraphics[width=\textwidth]{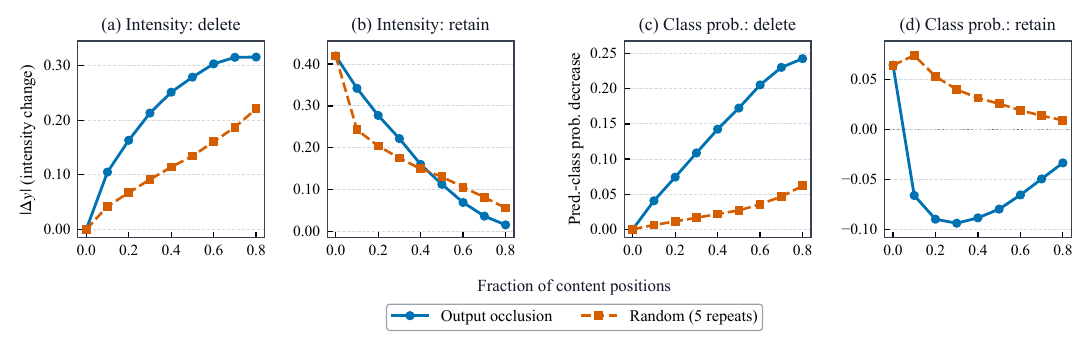}
\caption{Deletion and retention curves of the fused model on the 728 validation utterances: (a,b)~intensity change $|\Delta y|$ under top-ranked deletion and retention; (c,d)~predicted-class probability decrease under deletion and retention. Solid blue lines follow occlusion ranking; dashed orange lines average five random permutations.}
\label{fig:fidelity}
\end{figure*}

\subsection{Robustness to pre-encoding text-span deletion}
\label{sec:pre-encoding}
Because post-encoding masking can leave contextual traces of masked words in surviving tokens, we test a stricter protocol on validation mixed40: deleting characters covered by masked spans before encoding, re-encoding with frozen BERT and each LoRA expert, mapping surviving subwords to original positions, and keeping audio, vision, and masks fixed (subword-offset boundaries can split words). Fused Macro-F1 drops from 0.6394 to 0.6065 (MAE 0.5722 to 0.5930), while RM-Net core reaches 0.5813. Fusion still outperforms the core, but post-encoding masking overstates performance when words are truly absent; this check ran only on validation and was unused for selection. Separately, a video-grouped five-fold training-split check yields a mean two-view Macro-F1 of 0.6170 (SD 0.0073) for one adapter, measuring single-configuration stability rather than a nested five-seed estimate (Appendix~\ref{app:fused}).

\begin{table}[t]
\centering
\small
\renewcommand{\arraystretch}{1.08}
\setlength{\tabcolsep}{3pt}
\begin{tabular}{lcccc}
\toprule
& \multicolumn{2}{c}{\textbf{Intensity ($|\Delta y|$)}} & \multicolumn{2}{c}{\textbf{Class prob.\ ($\Delta p$)}}\\
\cmidrule(lr){2-3}\cmidrule(lr){4-5}
& Del.\ ($\uparrow$) & Ret.\ ($\downarrow$) & Del.\ ($\uparrow$) & Ret.\ (desc.)\\
\midrule
Selected & \textbf{0.2235} & 0.1796 & \textbf{0.1370} & $-$0.0645\\
\zrow
Random & 0.1136 & 0.1659 & 0.0244 & 0.0365\\
\midrule
\hirow
Paired diff. & \textbf{\tpos{0.1100}} & \tneg{0.0137} & \textbf{\tpos{0.1126}} & $-$0.1010\\
95\% CI low & 0.1010 & 0.0044 & 0.1081 & $-$0.1059\\
\zrow
95\% CI high & 0.1201 & 0.0238 & 0.1174 & $-$0.0963\\
\bottomrule
\end{tabular}
\caption{Deletion and retention AUCs for the fused model on 728 validation utterances (Del.: deletion; Ret.: retention; desc.: descriptive signed change). Intervals come from 500 paired bootstrap resamples over utterances (keeping each curve intact), without clustering by source video.}
\label{tab:fidelity}
\end{table}

\begin{figure*}[t]
\centering
\includegraphics[width=\textwidth]{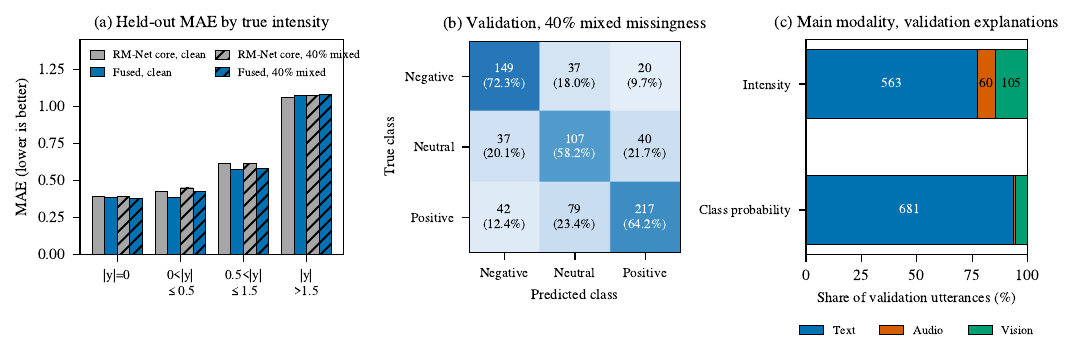}
\caption{Error diagnostics of the fused model. (a) Held-out MAE by true-intensity group (727 utterances) for the RM-Net core and fused model under clean and 40\% mixed-missing inputs. (b) Validation confusion matrix under 40\% mixed missingness (rows: true class; columns: predicted class; row-normalized percentages). (c) Dominant modality in validation explanations (728 utterances each): intensity---text 563, vision 105, audio 60; class probability---text 681, vision 40, audio 7.}
\label{fig:diagnostics}
\end{figure*}

\subsection{Dual-target explanations}
Table~\ref{tab:fidelity} and Figure~\ref{fig:fidelity} evaluate explanation fidelity on 728 validation utterances. Deleting top-ranked positions lowers predicted-class probability markedly faster than random deletion (AUC 0.137 vs.\ 0.024; paired diff.\ 0.113, CI $[0.108,0.117]$), and intensity deletion agrees (0.224 vs.\ 0.114). Retention does not establish sufficiency: top-ranked positions yield higher intensity retention AUC of $|\Delta y|$ than random (0.1796 vs.\ 0.1659), indicating larger average deviation from full-input intensity. Negative signed class retention AUC ($-$0.0645 vs.\ 0.0365) reflects an average probability increase across the integrated curve rather than guaranteed preservation per utterance or fraction; retention is therefore descriptive.

Shapley efficiency holds within $4.4\times10^{-16}$, and 12 recomputations (three unlabeled utterances across four conditions) match within $2.4\times10^{-7}$ for intensity and $2.7\times10^{-8}$ for probability with identical classes (reusing float32 text contexts while recomputing pooling, prediction, and fusion). First class flips along the selected order---censoring non-flipping utterances at $K+1$ for $K$ content positions---yield a median of 11 deleted positions versus 18 for random permutations (averaged per utterance over five permutations before taking the median); 272 of 728 utterances (37.4\%) never flip. The 357 validation utterances with opposite-sign text and vision class contributions show a 38.9\% error rate versus 32.4\% otherwise, useful for review triage without implying causation (Appendix~\ref{app:explanation}).

\subsection{Error diagnostics}
Figure~\ref{fig:diagnostics}(a) shows fusion lowers held-out MAE for neutral, weak, and moderate sentiment on both views, but raises strong-sentiment MAE from 1.063 to 1.071 (clean) and 1.074 to 1.077 (mixed40). In the mixed40 confusion matrix (b), 107 of 223 Neutral predictions are correct, while 37 are truly Negative and 79 Positive (precision 0.480, recall 0.582); Neutral is also weakest on clean inputs (Appendix~\ref{app:fused}). Panel~(c) shows text dominates 563 of 728 intensity attributions and 681 of 728 class attributions (vision: 105 and 40; audio: 60 and 7), reflecting model dependence rather than an absence of acoustic or visual cues.

\subsection{Unlabeled predictions}
On the 30 unlabeled utterances with zero spans (set A), fusion predicts 6 Negative, 13 Neutral, and 11 Positive; visible-text fractions span 0.50--1.00, with 27 utterances containing \texttt{[UNK]}. On the 20-utterance explanation set (set B), predictions are 8 Negative, 5 Neutral, and 7 Positive; text dominates all 20 class attributions and 16 intensity attributions (vision: 4). Lacking labels, we report only distributions and explanations (Appendix~\ref{app:predictions}).

\subsection{Auditable word-level alignment for raw clips}
\label{sec:alignment-summary}
A standalone pipeline extracts word-level features from 100 raw clips (37 videos) in the supplied CMU-MOSEI-derived package with transcripts and labels. CTC forced alignment over wav2vec~2.0 posteriors \citep{graves2006ctc,baevski2020wav2vec} supplies word boundaries, pairing WordPiece-averaged BERT vectors with 74 acoustic dimensions (10 ms hop) and 35 facial dimensions (25 Hz). Fixed thresholds on word/clip-mean posteriors and greedy WER pass 1,316 of 1,926 words (68.3\%) across 65 clips, and two independent runs match across all 500 files. Word-level grouping retains less acoustic variance than 0.25 s windows (0.249 vs.\ 0.312) but more than 0.5 s and 1 s windows (0.185 and 0.107)---a descriptive metric, not sentiment accuracy. Outputs never enter our predictive models (Appendix~\ref{app:alignment}, Figures~\ref{fig:alignment-example}--\ref{fig:alignment-quality}).

\section{Discussion}

Three bounded findings emerge. First, preserving indices and marking gaps enables aligned position-level auditing, though without a position-dropping control accuracy cannot be attributed to index retention alone. Second, fusing RM-Net with a LoRA BERT expert lifts the held-out two-view average---most strongly when text survives---while 60\% text masking narrows Macro-F1 gains and reverses them on T and TAV; isolating availability gating requires a fixed-weight control. Third, occlusion ranking removes positions that move predicted-class probability far more than random removal, whereas retention curves and text--vision conflicts remain descriptive diagnostics.

Deployment caveats persist: gains vary across conditions, and post-encoding masking overstates accuracy under genuine word loss. Validation-driven selection and prior held-out inspection make our figures estimates rather than a fresh blind test; accordingly, we present a transparent, auditable methodology rather than a state-of-the-art claim.

\section{Conclusion}

We introduced a reliability-aware multimodal sentiment framework pairing position-preserving RM-Net with a LoRA-adapted BERT text expert via deterministic visible-text gating. With all decisions frozen on validation, fusion improves held-out two-view Macro-F1 by 0.024 over the auxiliary-head baseline, peaking when lexical content remains intact. Exact Shapley attributions and single-position occlusion further localize influential inputs on the deployed predictor. Future work includes fresh blind evaluation, human explanation validation, probability calibration, and integrating word-level alignment into prediction.

\section*{Limitations}

\paragraph{Evaluation scope and missingness simulation.}
Several factors constrain the generality of our findings. Every modeling and fusion choice relied on the validation split, and the held-out split was inspected during earlier development; consequently, the held-out results are not a blind test and carry uncertainty from a single 727-utterance partition. Moreover, our main missingness protocols mask precomputed features after BERT context encoding, allowing surviving tokens to carry contextual leakage from masked words and thereby overstating performance under true word loss. Our stricter pre-encoding deletion check mitigates this concern only on the validation split under mixed40.

\paragraph{Modeling, weighting, and calibration.}
Fusion gains are uneven across conditions: heavy text masking degrades Macro-F1 relative to the baseline, including a 1.44-point drop when all three modalities lose 60\% of positions. Text-availability weighting follows a fixed rule selected on validation, and the learned class biases improve validation scores without improving held-out ranking. Furthermore, predicted probabilities are uncalibrated and should not be read as confidence estimates, nor was ensemble spread evaluated for uncertainty coverage.

\paragraph{Explanations, alignment, and transferability.}
Our explanations faithfully capture deployed-predictor input dependence under the stated interventions, not human emotional causation. The confidence intervals in Table~\ref{tab:fidelity} resample utterances rather than source videos, potentially understating uncertainty due to within-video correlation, and retention curves do not demonstrate evidence sufficiency. Cross-seed stability and parameter-randomization checks for the fused predictor's occlusion rankings remain future work, and no human evidence annotations were collected. Likewise, the standalone word-level alignment pipeline lacks human boundary ground truth, relies on heuristic time-quality thresholds, and does not feed the predictive models. Finally, our experiments cover a single English dataset with pre-extracted acoustic/visual features and a fixed text encoder; generalization to other languages, domains, or raw sensors, real-time physical-device latency, and full cumulative GPU compute were not evaluated.

\section*{Ethical Considerations}

\paragraph{Data privacy and misuse risks.}
CMU-MOSEI comprises online monologue videos of speakers. Our predictive experiments rely exclusively on precomputed numerical descriptors rather than raw imagery and make no attempt to identify speakers. The independent alignment experiment decodes 100 raw clips from 37 videos in the supplied CMU-MOSEI-derived package together with their transcripts, using face detection solely to extract per-frame numerical descriptors. Because the official CMU-MOSEI SDK distributes processed features rather than raw videos, we did not verify the supplied clips against their original YouTube sources. The separate code supplement excludes raw videos and strips transcript text from CSV evidence, whereas the EmoMosaic data repository includes source media and transcripts. All data were used solely for research. Multimodal sentiment analysis can be misused for surveillance, manipulation, or unfair screening, and strong average performance can mask disparities across speakers, languages, or expressive styles. Our explanations measure model input dependence---not a person's internal emotional state---and must not be used as evidence about individuals. Any downstream deployment should restrict use to consenting participants, audit subgroup performance, and keep humans in the loop for consequential decisions.

\paragraph{Licenses and redistribution.}
Our released code is distributed under the MIT license. The EmoMosaic data repository provides the four supplied, reorganized CMU-MOSEI-derived attachments---including available clips, transcripts, labels, and pre-extracted features---which represent a modified subset rather than the full CMU-MOSEI corpus. Our code license does not relicense third-party assets, whose original terms remain in force. Pretrained components (BERT, wav2vec~2.0, MediaPipe, and the Qwen backbones) are used under their original licenses and excluded from the code supplement, which ships only download URLs, revisions, and checksums. The released RM-Net and LoRA checkpoints were trained on CMU-MOSEI features and are provided strictly for research use.

\paragraph{Offensive content.}
Because the released dataset includes source video clips and transcripts, it may contain offensive language or otherwise objectionable content; no systematic offensive-content audit of the source data was conducted.

\paragraph{Environmental considerations.}
RM-Net checkpoints were trained on four consumer-grade RTX 3090 GPUs using BF16 mixed precision. The released core checkpoint is roughly 3 MB, and large pretrained backbones are excluded from distribution. Because cumulative GPU hours and energy consumption were not logged throughout development, we cannot report an exact carbon footprint, as noted in Section~\ref{sec:protocol} and Limitations.

\section*{Acknowledgments}

This work was supported in part by the Major Key Project of Pengcheng Laboratory under Grant PCL2025A12-3, and in part by the Science and Technology Planning Project of Guangdong Science and Technology Department under the Guangdong Key Laboratory of Advanced IntelliSense Technology under Grant 2019B121203006.\par
\noindent\textbf{Code and data availability.} Source code: \url{https://github.com/rudykon/EmoMosaic}; data repository: \url{https://huggingface.co/datasets/config-h/EmoMosaic}.

\AddToHookNext{shipout/after}{\AddToHookNext{shipout/after}{\balance}}
\bibliography{references}

\clearpage
\nobalance
\appendix
\renewcommand{\topfraction}{0.90}
\renewcommand{\bottomfraction}{0.85}
\renewcommand{\textfraction}{0.08}
\renewcommand{\floatpagefraction}{0.85}
\renewcommand{\dbltopfraction}{0.90}
\renewcommand{\dblfloatpagefraction}{0.85}
\setcounter{topnumber}{4}
\setcounter{bottomnumber}{3}
\setcounter{dbltopnumber}{2}
\setcounter{totalnumber}{6}
\setlength{\floatsep}{9pt plus 2pt minus 2pt}
\setlength{\textfloatsep}{11pt plus 2pt minus 2pt}
\setlength{\intextsep}{9pt plus 2pt minus 2pt}
\setlength{\dblfloatsep}{9pt plus 2pt minus 2pt}
\setlength{\dbltextfloatsep}{11pt plus 2pt minus 2pt}
\makeatletter
\setlength{\@fptop}{0pt plus 0.15fil}
\setlength{\@fpsep}{10pt plus 0.15fil}
\setlength{\@fpbot}{0pt plus 1fil}
\setlength{\@dblfptop}{0pt plus 0.15fil}
\setlength{\@dblfpsep}{10pt plus 0.15fil}
\setlength{\@dblfpbot}{0pt plus 1fil}
\makeatother

\input{appendix/app_model}
\input{appendix/app_data}
\input{appendix/app_core}
\input{appendix/app_fused}
\input{appendix/app_selection}
\input{appendix/app_alignment}
\input{appendix/app_explanation}
\input{appendix/app_predictions}

\begin{samepage}
\input{appendix/app_ai}
\par
\end{samepage}

\end{document}

%% file: appendix/app_model.tex
\section{Model Details}
\label{app:model}

This appendix provides the complete mathematical formulation corresponding to Section~\ref{sec:method}, following the same notation throughout.

\subsection{Observation masks and missingness operator}
\label{app:masks}

Let $\ell\le 50$ denote the number of valid positions in a modality, with validity indicator $p_i=\mathbf 1\{0\le i<\ell\}$ (valid positions include the special tokens of the text sequence). The observation indicator on raw features is
\begin{equation*}
o_i^{m,0}=p_i\,\mathbf 1\{\|x_i^m\|_0>0\}.
\end{equation*}
Given a synthetic missingness set $\mathcal M_m$ (a union of half-open spans), the damaged indicator is $o_i^{m,c}=o_i^{m,0}\mathbf 1\{i\notin\mathcal M_m\}$, and the unobserved fraction is $\rho_m^c=1-\frac1\ell\sum_i o_i^{m,c}$. Because observation status is computed from raw features prior to normalization, normalized features equal to zero are never misclassified as missing. Observed frames are z-scored using means and standard deviations computed over observed frames in the 3,395 training utterances, clipped to $[-5,5]$, and unobserved positions are reset to zero. The masking operator $\mathcal C_{\mathcal M}$ zeroes both features and indicators at masked indices without altering sequence length, preserving temporal alignment across all three modalities.

\subsection{Network equations}
\label{app:network}

Each modality $m\in\{t,a,v\}$ with raw input dimension $d_m\in\{768,74,35\}$ is projected to 128 dimensions. Unobserved positions are substituted with a learned \texttt{[MISS]} embedding $\mu_m$, and learned positional embeddings $\mathrm{PE}_i$ are added:
\begin{equation*}
\begin{aligned}
e_i^m&=o_i^m\bigl(P_m\bar x_i^m+b_m^{\mathrm{proj}}\bigr)\\
&\quad +(1-o_i^m)\mu_m+\mathrm{PE}_i,
\end{aligned}
\end{equation*}
where $P_m\in\mathbb R^{128\times d_m}$. Each modality is encoded by a two-layer, four-head Transformer with feed-forward width 256 and layer normalization; only padding positions are masked, keeping internal missing spans in place. A cross-modal layer allows each position of modality $m$ to attend to observed positions of the other two modalities, modulated by a learned distance bias
\begin{equation*}
b_m(i,j)=-\operatorname{softplus}(\beta_m)\,|i-j|/5.
\end{equation*}
When neither of the other modalities has any observed position, a zero-valued sentinel key is activated so attention remains well defined; this sentinel is visible only in that degenerate case.

Attention pooling aggregates strictly over observed positions:
\begin{equation*}
\begin{aligned}
u_i^m&=q_m^\top\tanh(U_m\widetilde h_i^m+b_m^{\mathrm{pool}}),\\
\alpha_i^m&=\operatorname{MSM}_i(u^m;o^m),\\
z_m&=\begin{cases}\sum_i\alpha_i^m\widetilde h_i^m,& r_m=1,\\ v_m^{\mathrm{absent}},& r_m=0,\end{cases}
\end{aligned}
\end{equation*}
where $r_m=\mathbf 1\{\sum_i o_i^m>0\}$ and $v_m^{\mathrm{absent}}$ is a learned absent vector. The masked softmax is defined as
\begin{equation*}
\operatorname{MSM}_j(u;v)=\begin{cases}\dfrac{v_j\exp(u_j)}{\sum_k v_k\exp(u_k)}, & \sum_k v_k>0,\\[4pt] 0, & \sum_k v_k=0.\end{cases}
\end{equation*}

The reliability gate combines each pooled modality vector with two observation descriptors and a reconstruction residual:
\begin{equation*}
\begin{aligned}
s_j&=g_j([z_j;\rho_j;n_j^{\mathrm{seg}}/\ell;\epsilon_j]),\\
w_m&=\operatorname{MSM}_m\!\bigl((s_j)_{j\in\{t,a,v\}};r\bigr),
\end{aligned}
\end{equation*}
where $n_j^{\mathrm{seg}}$ counts contiguous unobserved segments within the valid range. The residual is evaluated exclusively on observed positions via a reconstruction head:
\begin{equation*}
\begin{aligned}
R_m&=\frac{\sum_i o_i^m\|\hat x_i^m-\bar x_i^m\|_2^2}
              {d_m\max(1,\sum_i o_i^m)},\\
\epsilon_m&=\frac{1}{25}\min\{25,\operatorname{sg}(R_m)\},
\end{aligned}
\end{equation*}
where $\operatorname{sg}$ denotes the stop-gradient operator and $\epsilon_m=0$ when $r_m=0$. This residual acts as an internal reliability proxy rather than a calibrated predictive variance.

The resulting intensity estimate is additive and bounded:
\begin{equation*}
\begin{gathered}
\hat y_m=2.7\tanh\bigl(a_m(z_m)\bigr),\\
\hat y_{\mathrm{int}}=\begin{cases}0.3\tanh\bigl(a_{\mathrm{int}}(z)\bigr),& \sum_m r_m>0,\\ 3\tanh(\eta),& \sum_m r_m=0,\end{cases}\\
\hat y=\sum_m w_m\hat y_m+\hat y_{\mathrm{int}},
\end{gathered}
\end{equation*}
with $z=\sum_m w_m z_m$ and a learned empty-set prior $\eta$. Each modality's exact contribution is $c_m=w_m\hat y_m$, yielding an exact additive decomposition of $\hat y$. Across the five-seed ensemble, we average per-seed contributions, $\frac15\sum_s w_m^{(s)}\hat y_m^{(s)}$, rather than multiplying averaged gates by averaged branch outputs. An auxiliary classifier maps $z$ to three class logits, and ensemble class probabilities are averaged after applying softmax per seed. The core network totals 1,510,206 parameters.

\subsection{Training objective and masking schedule}
\label{app:training}

Within each mini-batch, the clean and damaged views share parameters under the objective
\begin{equation*}
\begin{aligned}
\mathcal L&=\mathcal L_{\mathrm{pred}}+0.3\mathcal L_{u}+0.3\mathcal L_{\mathrm{CE}}\\
&\quad+0.5\mathcal L_{\mathrm{rec}}+0.5\mathcal L_{\mathrm{con}}+0.1\mathcal L_{\mathrm{int}}.
\end{aligned}
\end{equation*}
Here, $\mathcal L_{\mathrm{pred}}=\frac1B\sum_n(|\hat y_n^c-y_n|+|\hat y_n^{\mathrm{clean}}-y_n|)$ computes the L1 intensity error on both views; $\mathcal L_u$ applies the same L1 loss to every modality that remains observed in the damaged view, normalized by the count of surviving modalities; and $\mathcal L_{\mathrm{CE}}$ is inverse-class-frequency-weighted cross-entropy across both views. Among the structural regularizers, $\mathcal L_{\mathrm{rec}}$ applies a smooth L1 penalty strictly to positions observed in the clean view and newly masked in the damaged view; $\mathcal L_{\mathrm{con}}$ aligns the damaged fused representation and intensity with stop-gradient clean targets; and $\mathcal L_{\mathrm{int}}$ penalizes the magnitude of $\hat y_{\mathrm{int}}$ in the damaged view.

Masking augmentation follows a fixed curriculum. Each sample remains unaugmented with probability 0.2; otherwise, 1, 2, or 3 modalities are masked with probabilities 0.5, 0.35, and 0.15, respectively. Each masked modality draws a missing ratio $r$ from $0.7\,U(0.1,0.6)+0.25\,U(0.6,0.95)+0.05\,\delta(1)$, applied as one or two contiguous spans (damaged modalities share span coordinates with probability 0.3). Span lengths equal $\lceil r\ell\rceil$ and may cover special tokens. Over the first 30\% of training, the upper bound on $r$ increases linearly from 0.3 to 1. Separately, augmented samples mask all three modalities with probability 0.01 to supervise the empty-set output.

Table~\ref{tab:hyper} summarizes all optimization and search settings. RM-Net checkpoints were selected using the validation criterion $J=r-\mathrm{MAE}+\mathrm{Macro\text{-}F}_1$, where $r$ is Pearson correlation; the clean condition and the mean over the 20\%, 40\%, and 60\% missing-rate conditions receive equal weight (0.5 each), and the intensity cut-points for Macro-F1 are re-tuned on validation predictions at every epoch. Across the five seeds, the selected checkpoints fell between epochs 8 and 11 (1-based). The LoRA text expert was selected by mean Macro-F1 across the validation clean and mixed40 views.

\begin{table*}[t]
\centering
\small
\renewcommand{\arraystretch}{1.08}
\begin{tabularx}{\textwidth}{@{}lY@{}}
\toprule
\textbf{Setting} & \textbf{Value}\\
\midrule
RM-Net optimizer & AdamW \citep{loshchilov2019adamw}, learning rate $10^{-4}$, weight decay $10^{-3}$\\
\zrow
RM-Net schedule & up to 60 epochs, patience 10, batch 64, 5\% warm-up, cosine decay\\
RM-Net regularization & dropout 0.25, input dropout 0.1, gradient clipping 1.0\\
\zrow
RM-Net precision & BF16 mixed precision\\
\midrule
Seeds & 17, 29, 43, 71, 101 (core and text expert)\\
\zrow
Text expert & rank 8, $\alpha=16$, query and value projections of all 12 layers; batch 32\\
Text expert optimizer & AdamW, weight decay 0.01; adapter learning rate $10^{-4}$, head learning rate $3\times10^{-4}$\\
\zrow
Text expert schedule & up to 10 epochs, patience 3\\
\midrule
Fusion search & fusion weights on $[0,1]$ (step 0.1); exponents $\{1,2,4\}$; class bias on $[-1,1]$ (step 0.1), refined with step 0.025\\
\zrow
Mixed40 placement seed & 20260923\\
\bottomrule
\end{tabularx}
\caption{Training and evaluation settings. Fusion search used only the validation split; the held-out split had been inspected during earlier development.}
\label{tab:hyper}
\end{table*}

\subsection{Text expert and fusion}
\label{app:fusion}

The text expert freezes BERT-base and inserts rank-8 LoRA adapters ($\alpha=16$) into the query and value projections of all 12 layers, adding roughly 0.29M trainable adapter parameters (excluding task heads). Mean pooling aggregates over visible positions that are not \texttt{[UNK]}, including visible special tokens, and outputs a zero vector if no usable position remains. The classification and intensity heads share one 128-unit hidden layer followed by separate linear output projections. The expert minimizes class-weighted cross-entropy across clean and damaged views plus $0.3$ times an L1 intensity loss. Fusion equations and frozen constants follow Section~\ref{sec:method}, with fusion weights, exponents, and class biases chosen on validation over the grids in Table~\ref{tab:hyper}.

Because both classification and intensity fusion depend on the same visible-text fraction $f$, their expert weights remain coupled as $w=0.8f^2$ and $v=0.6f^2$. Across unlabeled set A, $f$ spans 0.50 to 1.00 (Appendix~\ref{app:predictions}).

%% file: appendix/app_data.tex
\section{Data Statistics and Input Checks}
\label{app:data}

\subsection{Split statistics}
\label{app:splits}

Table~\ref{tab:splits} summarizes the aligned data release across all three splits. Median lengths measure the number of valid positions per utterance. Zero-frame ratios denote the fraction of valid positions whose raw audio or visual vector is identically zero; because structural zeros, natural silence or absent faces, and pre-existing missing spans cannot be distinguished in the release, these sources remain combined.

\begin{table*}[tp]
\centering
\small
\renewcommand{\arraystretch}{1.08}
\setlength{\tabcolsep}{7pt}
\begin{tabular}{lrrrrrr}
\toprule
& & \multicolumn{3}{c}{\textbf{Class distribution}} & & \\
\cmidrule(lr){3-5}
\textbf{Split} & \textbf{$N$} & \tneg{Neg} & Neu & \tpos{Pos} & \shortstack{\textbf{Median}\\\textbf{length}} & \shortstack{\textbf{Zero audio /}\\\textbf{visual frames}}\\
\midrule
Train & \textbf{3,395} & 967 & 758 & 1,670 & 22.0 & 0.082 / 0.132\\
\zrow
Valid & \textbf{728} & 206 & 184 & 338 & 23.5 & 0.081 / 0.131\\
Test & \textbf{727} & 207 & 158 & 362 & 23.0 & 0.079 / 0.135\\
\bottomrule
\end{tabular}
\caption{Statistics of the aligned release (50 positions per modality). Neutral corresponds strictly to $y=0$ and comprises 22.3\% of the training split.}
\label{tab:splits}
\end{table*}

The validation split spans 239 source videos. Two properties of the release directly shape our processing: first, the 50 aligned indices correspond to text token positions rather than words or uniform time intervals; second, padding positions in the precomputed text features are non-zero, so valid lengths must be read from the attention mask rather than inferred from the text vectors.

\subsection{Re-extraction of text features}
\label{app:reextraction}

Text features were re-extracted across all splits and unlabeled sets using the frozen uncased BERT-base encoder at a fixed revision so that the RM-Net text branch and the LoRA text expert consume identical inputs. On observed positions, re-extracted vectors match the supplied text features within a mean absolute difference on the order of $3\times10^{-4}$, consistent with half-precision storage.

\subsection{Availability in the unlabeled sets}
\label{app:availability}

In unlabeled set A (30 utterances), 27 utterances contain at least one \texttt{[UNK]} token, totaling 131 positions. Although these positions carry non-zero BERT vectors, they are excluded from the numerator of the visible-text fraction under the rule in Section~\ref{sec:method}, while \texttt{[CLS]} and \texttt{[SEP]} are excluded from both numerator and denominator. As a result, the visible-text fraction ranges from 0.50 to 1.00 across set A. Because the release does not distinguish structural, natural, and artificially imposed zero spans in audio or vision, the model treats a position as observed only when its raw feature vector is non-zero.

%% file: appendix/app_core.tex
\section{Robustness Scans of the Core Network}
\label{app:core}

This appendix presents controlled robustness scans of the standalone RM-Net core prior to fusion with the text expert. Classification curves in these scans use the historical intensity-threshold rule on the validation split and therefore do not reflect the fused classifier. In contrast, intensity MAE is independent of that classification rule and illustrates how the multimodal core degrades under each missingness pattern. All scans employ the same contiguous-span masking operator used in the main experiments.

\subsection{Missing fraction and modality subset}
\label{app:core-fraction}

\begin{figure*}[tp]
\centering
\includegraphics[width=0.92\textwidth]{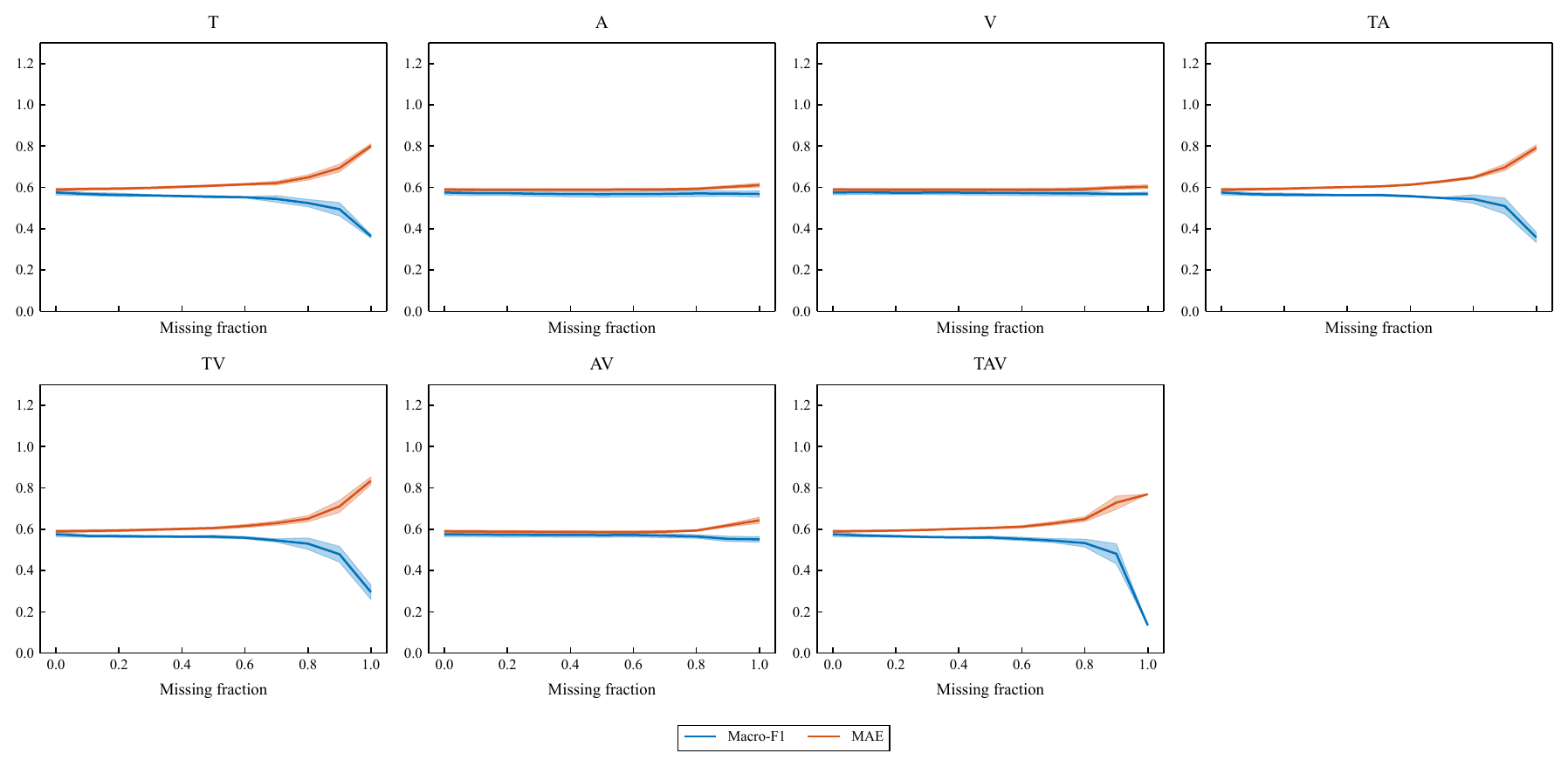}
\caption{RM-Net core performance under increasing missingness on the validation split. Each panel corresponds to one damaged-modality subset; blue curves show Macro-F1 (historical thresholds) and orange curves show MAE. Shading indicates the standard deviation across five training seeds after averaging over span placements within each seed.}
\label{fig:core-fraction}
\end{figure*}

Figure~\ref{fig:core-fraction} sweeps the missing fraction from 0 to 1 in increments of 0.1 across all seven damaged-modality subsets, using five training seeds and five span placements. Text missingness dominates degradation: MAE reaches roughly 0.648 at 80\% text loss and rises to 0.801 when text is completely absent, whereas complete audio or visual loss alone yields MAEs of roughly 0.611 and 0.604. When all three modalities are absent, predictions fall back to the learned empty-set prior. These scans characterize local degradation and do not imply robustness under total information loss.

\subsection{Span position}
\label{app:core-position}

\begin{figure*}[tp]
\centering
\includegraphics[width=0.85\textwidth]{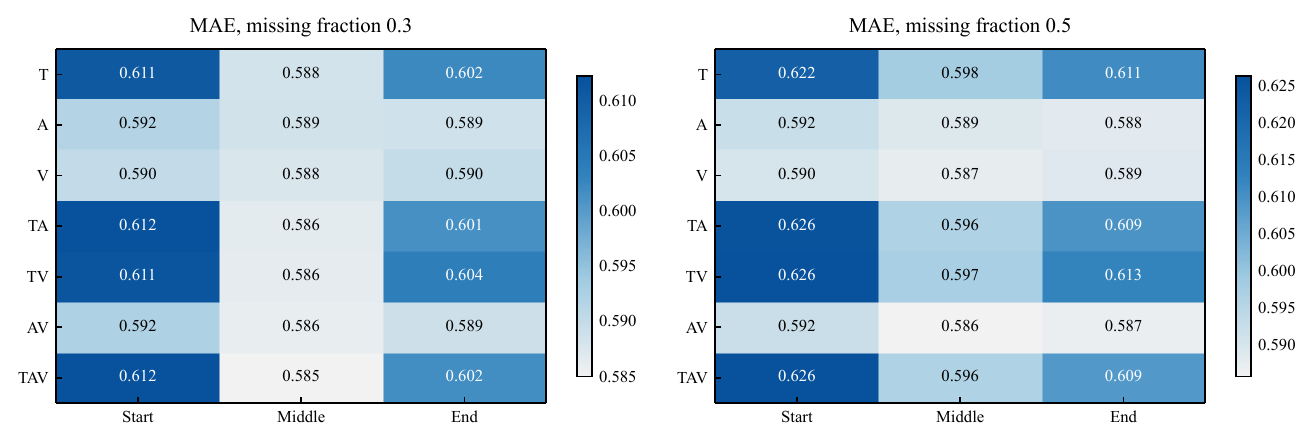}
\caption{MAE of the RM-Net core by damaged-modality subset (rows) and span position (columns) at missing fractions 0.3 (left) and 0.5 (right). Color scales differ between the two panels.}
\label{fig:core-position}
\end{figure*}

Figure~\ref{fig:core-position} compares missing spans placed at the start, middle, and end of the admissible range at missing fractions 0.3 and 0.5. Averaged across both rates, text MAE is 0.616 at the start, 0.593 in the middle, and 0.607 at the end, making the utterance start the most sensitive location. Because start-aligned spans overlap both the \texttt{[CLS]} token and initial content words, this scan cannot disentangle the two effects. Positional differences for audio and vision remain negligible.

\subsection{Span length and reconstruction}
\label{app:core-length}

\begin{figure*}[tp]
\centering
\includegraphics[width=0.92\textwidth]{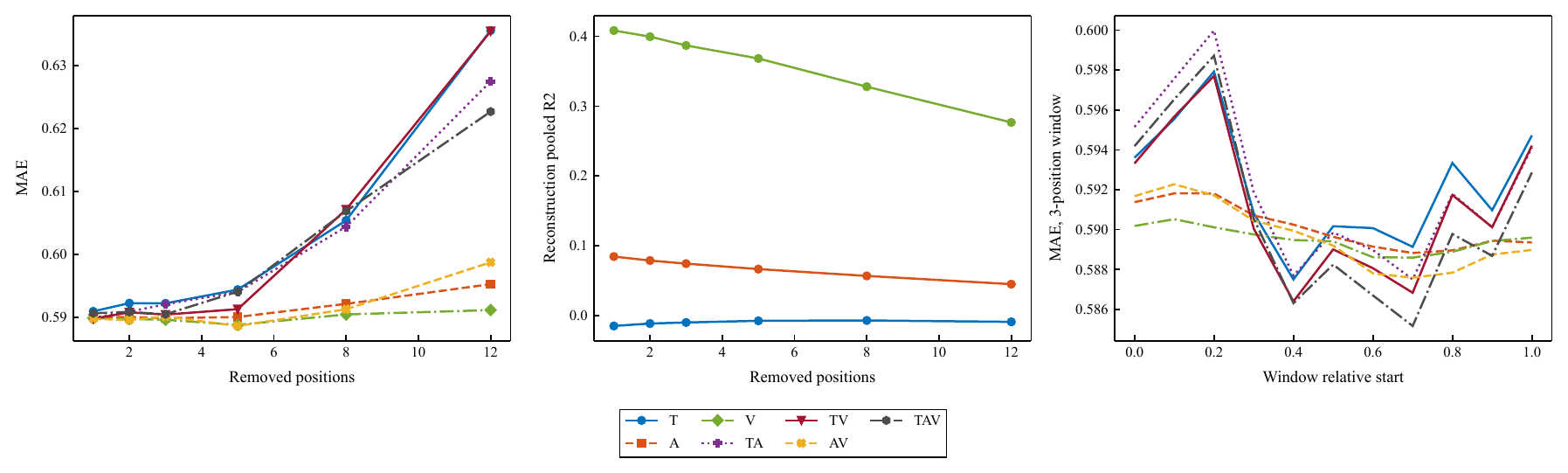}
\caption{RM-Net core sensitivity to span length and placement. Left: MAE versus the number of removed positions by modality subset. Middle: reconstruction $R^2$ on removed positions under single-modality masking. Right: MAE of a three-position window as its relative starting offset varies.}
\label{fig:core-length}
\end{figure*}

Figure~\ref{fig:core-length} masks 1, 2, 3, 5, 8, or 12 valid positions. On the removed positions, visual reconstruction $R^2$ declines from 0.408 to 0.277, audio falls from 0.084 to 0.045, and text stays between $-0.015$ and $-0.007$. Thus, cross-modal feature reconstruction is informative for vision, weak for audio, and no better than a constant predictor for text. The right panel slides a three-position window across the admissible range; MAE fluctuates by roughly 0.015, indicating modest sensitivity to placement for short gaps.

Two additional diagnostics warrant note. First, at 40\% text missingness, masking one contiguous span yields an MAE of 0.603 versus 0.599 for four disjoint spans (not tested for significance). Second, as text missingness rises from 10\% to 90\%, the core's mean text gate weight increases slightly from 0.531 to 0.548 rather than decreasing monotonically, confirming that the gate operates as an internal learned weight rather than a calibrated reliability estimate.

%% file: appendix/app_fused.tex
\section{Fused-Model Missingness and Validation Diagnostics}
\label{app:fused}

This appendix expands the fused-model diagnostics summarized in Section~\ref{sec:results}. Because the validation split guided model development, all validation metrics reported here characterize the frozen predictor on development data rather than a blind test.

\subsection{Per-utterance changes under mixed missingness}
\label{app:flips}

Across the 728 validation utterances, 447 are classified correctly under both clean and 40\% mixed-missing inputs, while 233 are misclassified under both. Missingness flips 22 correct predictions to incorrect and 26 incorrect predictions to correct, showing that aggregate Macro-F1 shifts mask partially offsetting per-utterance changes rather than a uniform degradation. Table~\ref{tab:flip-examples} presents one illustrative flip per true class---specifically, the flipped utterance exhibiting the largest increase in absolute intensity error within each class. These examples document failure modes and were not used for selection. Spans use 0-based half-open indices, and ``$\varnothing$'' denotes no additional mask.

\begin{table*}[tp]
\centering
\small
\renewcommand{\arraystretch}{1.08}
\begin{tabular}{lcccc}
\toprule
\textbf{Case} & \shortstack{\textbf{True polarity /}\\\textbf{intensity}} & \shortstack{\textbf{Polarity}\\\textbf{change}} & \shortstack{\textbf{Intensity}\\\textbf{change}} & \textbf{Text span} \\
\midrule
\textbf{V1} (\texttt{Vuay0mVKkik\$\_\$3}) & \tneg{Neg} / $-2.0$ & \tneg{Neg}$\to$\tpos{Pos} & $-0.201\to0.485$ & $[2,14)$\\
\zrow
\textbf{V2} (\texttt{fWOIAxzBQFY\$\_\$7}) & Neu / $0.0$ & Neu$\to$\tpos{Pos} & $0.453\to0.805$ & $[12,21)$\\
\textbf{V3} (\texttt{128059\$\_\$6}) & \tpos{Pos} / $1.0$ & \tpos{Pos}$\to$\tneg{Neg} & $0.039\to-0.729$ & $[15,35)$\\
\bottomrule
\end{tabular}
\caption{Representative validation flips under mixed missingness. Audio and visual spans are $[4,16)$ and $[15,27)$ for V1, $[6,15)$ and $[3,12)$ for V2, and $[30,50)$ and $\varnothing$ for V3.}
\label{tab:flip-examples}
\end{table*}

In V1, which describes difficulty obtaining educational resources, the masked text span covers the phrase ``really difficult for me''; removing that span alongside separate audio and visual spans flips the prediction from Negative to Positive. V2 is a factual statement whose Neutral prediction shifts to Positive when a span near the text tail is masked. V3 expresses a positive evaluation; masking that span together with the audio tail flips the prediction from Positive to Negative. While these cases illustrate how local cue loss can reverse decisions, multiple modalities are perturbed concurrently, preventing single-cause attribution.

\subsection{Position and length groups}
\label{app:position-length}

Table~\ref{tab:position-length} reports the paired validation MAE increase (masked minus clean) across span positions and lengths. Length groups pool the 20\%, 40\%, and 60\% missingness conditions (2,184 records per modality across 728 utterances), so records from the same utterance are correlated. Text masking at the head, middle, and tail of the admissible range raises MAE by 0.0214, 0.0165, and 0.0287, respectively, whereas audio and visual masking alter MAE by at most 0.0022. Longer text spans coincide with steeper MAE increases, but because missing rate and utterance length vary across groups, this trend does not isolate absolute span length.

\begin{table*}[tp]
\centering
\small
\renewcommand{\arraystretch}{1.08}
\begin{tabular}{lcccc}
\toprule
\textbf{Grouping} & \textbf{Group 1} & \textbf{Group 2} & \textbf{Group 3} & \textbf{Group 4}\\
\midrule
Span position: head / middle / tail (40\%) & 0.0214 & 0.0165 & \cneg\textbf{\tneg{0.0287}} & --\\
\midrule
Span length: 1--4 / 5--8 / 9--12 / $\ge$13 (text) & 0.0034 & 0.0141 & 0.0192 & \cneg\textbf{\tneg{0.0343}}\\
\zrow
Records per length group & 352 & 658 & 516 & 658\\
\bottomrule
\end{tabular}
\caption{Paired validation MAE increase under text masking (masked minus clean; higher values indicate larger degradation). Red \textbf{bold} cells mark the largest error increase in each grouping. Position rows use the 40\% condition; length rows pool the 20\%, 40\%, and 60\% conditions.}
\label{tab:position-length}
\end{table*}

\subsection{Intensity groups and per-class behavior}
\label{app:groups}

Table~\ref{tab:groups} breaks down validation MAE for the fused model by true sentiment intensity. Under mixed40, MAE rises noticeably for the moderate group (0.568 to 0.583), while strong sentiment ($|y|>1.5$) remains the dominant error source in both conditions. The neutral group corresponds strictly to $|y|=0$.

\begin{table}[tp]
\centering
\small
\renewcommand{\arraystretch}{1.08}
\setlength{\tabcolsep}{4pt}
\begin{tabularx}{\linewidth}{@{}Yrcc@{}}
\toprule
\textbf{Group ($|y|$)} & \textbf{$n$} & \shortstack{\textbf{Clean}\\\textbf{MAE ($\downarrow$)}} & \shortstack{\textbf{Mixed40}\\\textbf{MAE ($\downarrow$)}}\\
\midrule
$0$ (Neutral) & 184 & 0.352 & 0.348\\
\zrow
$0<|y|\le0.5$ (weak) & 147 & 0.391 & 0.399\\
$0.5<|y|\le1.5$ (moderate) & 267 & 0.568 & 0.583\\
\zrow
$|y|>1.5$ (strong) & 130 & \cneg\textbf{\tneg{1.057}} & \cneg\textbf{\tneg{1.064}}\\
\bottomrule
\end{tabularx}
\caption{Validation MAE of the fused model stratified by true-intensity group. Red shading highlights strong sentiment as the dominant error source.}
\label{tab:groups}
\end{table}

Table~\ref{tab:perclass} reports per-class precision, recall, and F1 on the clean validation split. Neutral is the weakest class across all three metrics, consistent with its mixed40 precision of 0.480 and recall of 0.582 (Figure~\ref{fig:diagnostics}).

\begin{table}[tp]
\centering
\small
\renewcommand{\arraystretch}{1.08}
\setlength{\tabcolsep}{4pt}
\begin{tabular}{@{}lccc@{}}
\toprule
\textbf{Class} & \shortstack{\textbf{Precision}\\\textbf{(\%, $\uparrow$)}} & \shortstack{\textbf{Recall}\\\textbf{(\%, $\uparrow$)}} & \shortstack{\textbf{F1}\\\textbf{(\%, $\uparrow$)}}\\
\midrule
\tneg{Negative} & 65.6 & \cbest\textbf{72.3} & 68.8\\
\zrow
Neutral & \cneg\textbf{\tneg{46.3}} & \cneg\textbf{\tneg{57.1}} & \cneg\textbf{\tneg{51.1}}\\
\tpos{Positive} & \cbest\textbf{78.5} & 63.6 & \cbest\textbf{70.3}\\
\bottomrule
\end{tabular}
\caption{Per-class classification metrics of the fused model on the clean validation split. Blue \textbf{bold} cells mark per-column highs; red \textbf{bold} cells highlight Neutral as the weakest class.}
\label{tab:perclass}
\end{table}

\subsection{Replay of the unlabeled-set zero spans}
\label{app:replay}

\begin{figure*}[tp]
\centering
\includegraphics[width=0.92\textwidth]{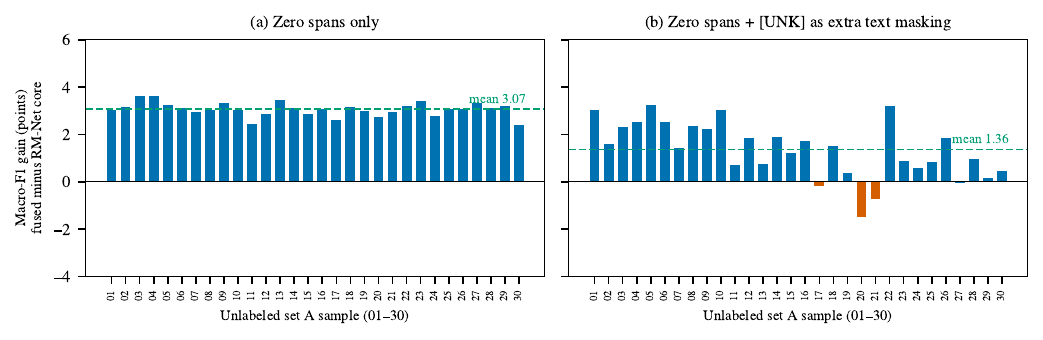}
\caption{Held-out replay of the unlabeled set A zero spans. Bars show the Macro-F1 gain of the fused model over the RM-Net core for each pattern (samples 01--30). Panel~(b) additionally masks the \texttt{[UNK]} positions of set A. Dashed horizontal lines mark the mean gain.}
\label{fig:replay}
\end{figure*}

To test naturally occurring missingness layouts, we replayed the 30 zero-span patterns from unlabeled set A onto all 727 held-out utterances by mapping relative span coordinates to each utterance's valid length. Figure~\ref{fig:replay} plots the resulting Macro-F1 gain of the fused model over the RM-Net core. Replaying only zero spans (panel~a) improves Macro-F1 across all 30 patterns (mean gain +3.07 points). When \texttt{[UNK]} positions from set A are additionally masked as missing text (panel~b), the mean gain falls to +1.36 points, with patterns 17, 20, and 21 turning negative. MAE improves across every pattern in both panels. These replays offer a condition-level view of the same behavior, though the 30 patterns are not independent samples.

\subsection{Pre-encoding text-span deletion and fold-level stability}
\label{app:strict}

In the pre-encoding deletion check on the validation mixed40 condition, characters covered by each masked text span are deleted prior to encoding; the remaining text is re-encoded with frozen BERT and each LoRA expert, surviving subwords are mapped back to their original positions, and audio, vision, and masks remain unchanged. Under this stricter protocol, fused Macro-F1 declines from 0.6394 to 0.6065 and MAE rises from 0.5722 to 0.5930, while the RM-Net core achieves 0.5813 Macro-F1. Because span boundaries follow subword offsets, a boundary can split an individual word, making this check pre-encoding span deletion rather than strictly whole-word deletion. This check was never used for model selection.

Separately, a video-grouped five-fold check was executed inside the training split, training one LoRA adapter per fold from public BERT weights. Across the five folds, two-view Macro-F1 averages $0.6170$ with a standard deviation of $0.0073$. This measures the fold-to-fold stability of a single adapter configuration rather than providing a nested estimate for the five-seed fused ensemble.

%% file: appendix/app_selection.tex
\section{Model Selection and Frozen-Representation Comparisons}
\label{app:selection}

The deployed fused predictor was selected from a structured candidate pool on the validation split. This appendix documents the full candidate set, the pre-registered replacement rule, and the non-adopted alternatives so that every selection decision remains auditable.

\subsection{Candidate workflow after the frozen model}
\label{app:workflow}

After freezing the final model, we evaluated nine post-freeze workflow candidates: the frozen baseline plus eight alternatives comprising two class-bias shrinkage variants (scaling bias by 0.5 or 0), three quality-adaptive fusion variants, and three quality-adaptive fusion variants augmented with a restricted review step. The review step triggers only when text is present and either the RM-Net and LoRA predicted classes disagree or the top two fused probabilities differ by less than 0.20, adding at most two extra RM-Net core calls (audio-visual only or text only). Adaptive fusion uses 27 diagnostic features, while the review variant uses 49; learned variants sweep L2 regularization over $\{0.01, 0.1, 1\}$ and are fitted on out-of-fold predictions from five video-grouped training folds.

A pre-specified replacement rule required any candidate to satisfy three criteria simultaneously on validation: (i) two-view mean Macro-F1 increases by at least 0.003, (ii) neither view's Macro-F1 decreases, and (iii) neither view's MAE increases by more than 0.01. Table~\ref{tab:workflow} reports the top validation configuration within each family.

\begin{table*}[tp]
\centering
\small
\renewcommand{\arraystretch}{1.08}
\setlength{\tabcolsep}{4pt}
\begin{tabularx}{\linewidth}{@{}Ylcccc@{}}
\toprule
\textbf{Family} & \textbf{Configuration} & \shortstack{\textbf{Val avg}\\\textbf{($\uparrow$)}} & \shortstack{\textbf{Test avg}\\\textbf{($\uparrow$)}} & \shortstack{\textbf{Test $\Delta$}\\\textbf{vs.\ frozen}} & \textbf{Adopted}\\
\midrule
\hirow
\fstar\;\textbf{Frozen RM + LoRA fusion} & \texttt{baseline} & \textbf{0.636652} & 0.644564 & 0 & \cmark\;\textbf{Yes}\\
\midrule
Class-bias shrinkage & \texttt{bias\_shrink\_0.5} & 0.629462 & 0.646465 & \tpos{+0.001902} & \xmark\;No\\
\zrow
Quality-adaptive fusion & \texttt{adaptive\_l2\_0.1} & 0.614761 & 0.650359 & \tpos{+0.005795} & \xmark\;No\\
Quality fusion + restricted review & \texttt{review\_l2\_0.01} & 0.626297 & \textbf{0.657514} & \tpos{+0.012950} & \xmark\;No\\
\bottomrule
\end{tabularx}
\caption{Post-freeze workflow candidates. Averages combine clean and 40\% mixed-missing Macro-F1. \fstar{} marks the retained frozen model (\cmark); none of the alternatives (\xmark) met the validation replacement rule. Held-out columns are shown for transparency; they were inspected earlier and were not used to choose among candidates.}
\label{tab:workflow}
\end{table*}

Although the best review candidate posts a held-out Macro-F1 difference of +0.0130 (95\% video-cluster bootstrap CI $[-0.0043,\,0.0303]$; two-sided permutation $p=0.163$, uncorrected for multiple comparisons), its validation average trails the frozen model and thus fails the replacement rule. Across the 21-condition missingness grid, it averages +0.0146 Macro-F1 (12 wins, 0 ties, 9 losses), triggering review on 31.9\% (clean) and 34.3\% (mixed40) of held-out utterances (+0.64 and +0.69 extra calls per utterance). Accordingly, we retain the frozen model and record the review workflow strictly as a documented alternative.

\subsection{Interface checks}
\label{app:interface}

An offline verification suite executed 156 raw-input inferences across four configurations: all 30 unlabeled set A utterances plus three boundary utterances each under mixed missingness, all-text-missing, and all-modalities-missing conditions. Predicted class labels matched cached-feature inference in every case, and all probability, intensity, and core-contribution identities held. This check validates software interface fidelity rather than accuracy on unlabeled inputs.

\subsection{All candidates}
\label{app:candidates}

Table~\ref{tab:candidates} lists all 29 frozen-representation and fusion candidates alongside their validation and held-out metrics. Two findings stand out. First, the final configuration achieves the highest validation average across the entire grid, leading the strongest frozen-representation fusion (RM-Net (CE 1, 15 seeds) + BERT probe fusion + bias) by 0.0003. Second, several alternatives lead on single held-out views---such as the standalone LoRA expert on mixed40 and unbiased fusion on clean inputs---without establishing a consistent ranking, underscoring why selection follows the fixed validation rule rather than held-out inspection.

The Qwen baselines attach linear probes to pooled hidden states from frozen Qwen2.5-1.5B, Qwen3.5-0.8B, Qwen3.5-2B, and Qwen3.5-4B encoders, selecting probe layer and regularization on the training and validation splits. None of the Qwen3.5 probes met the preset validation fusion threshold, and every Qwen probe trails the fused model in held-out Macro-F1. SVD baselines pair singular-value features with logistic regression and serve strictly as lightweight references.

\input{appendix/tab_candidates}

\subsection{Grouped stability of the text expert}
\label{app:stability}

The training-split stability check in Appendix~\ref{app:fused} trains one LoRA adapter per video-grouped fold. We summarize its fold-level Macro-F1 strictly as a mean and standard deviation, since single-adapter training folds do not constitute independent estimates of the five-seed fused model.

%% file: appendix/tab_candidates.tex
\begin{table*}[tp]
\centering
\small
\renewcommand{\arraystretch}{1.04}
\setlength{\tabcolsep}{3pt}
\begin{tabularx}{\linewidth}{@{}Y*{7}{c}@{}}
\toprule
& \multicolumn{3}{c}{\textbf{Val Macro-F1 ($\uparrow$)}} & \multicolumn{2}{c}{\textbf{Test Macro-F1 ($\uparrow$)}} & \multicolumn{2}{c}{\textbf{Test MAE ($\downarrow$)}}\\
\cmidrule(lr){2-4}\cmidrule(lr){5-6}\cmidrule(lr){7-8}
\textbf{Configuration} & Clean & Mixed & Avg & Clean & Mixed & Clean & Mixed\\
\midrule
RM-Net core (auxiliary head) & 0.6144 & 0.6057 & 0.6101 & 0.6237 & 0.6178 & 0.6344 & 0.6379\\
\zrow
RM-Net core + class bias & 0.6169 & 0.6177 & 0.6173 & 0.6185 & 0.6189 & 0.6344 & 0.6379\\
BERT last-layer probe & 0.5976 & 0.5935 & 0.5956 & 0.6280 & 0.6118 & 0.6870 & 0.6969\\
\zrow
BERT multi-layer probe & 0.6118 & 0.6007 & 0.6063 & 0.6315 & 0.6371 & 0.6950 & 0.7019\\
BERT + SVD probe & 0.6125 & 0.5930 & 0.6027 & 0.6090 & 0.6161 & 0.7004 & 0.7032\\
\zrow
BERT scalar-mix probe (5 seeds) & 0.6074 & 0.6165 & 0.6119 & 0.6209 & 0.6261 & 0.6436 & 0.6432\\
BERT scalar-mix probe (5 seeds) + bias & 0.6179 & 0.6276 & 0.6227 & 0.6150 & 0.6152 & 0.6436 & 0.6432\\
\zrow
SVD + logistic regression, all modalities + bias & 0.6207 & 0.6146 & 0.6176 & 0.6213 & 0.6136 & 0.6344 & 0.6379\\
SVD + logistic regression, text only + bias & 0.6163 & 0.6087 & 0.6125 & 0.6309 & 0.6192 & 0.6344 & 0.6379\\
\zrow
Qwen2.5-1.5B probe & 0.5962 & 0.5791 & 0.5876 & 0.5853 & 0.5777 & 0.6845 & 0.7794\\
Qwen3.5-0.8B probe & 0.5788 & 0.5704 & 0.5746 & 0.5965 & 0.5676 & 0.7448 & 0.7665\\
\zrow
Qwen3.5-2B probe & 0.6021 & 0.5810 & 0.5915 & 0.6192 & 0.6010 & 0.7100 & 0.7727\\
Qwen3.5-4B probe & 0.6028 & 0.5932 & 0.5980 & 0.6027 & 0.5867 & 0.7046 & 0.7479\\
\zrow
LoRA text expert only & 0.6175 & 0.6269 & 0.6222 & \underline{0.6578} & \textbf{0.6524} & 0.6171 & 0.6223\\
\midrule
RM-Net, CE weight 0.3, 15 seeds, bias & 0.6153 & 0.6216 & 0.6185 & 0.6126 & 0.6144 & 0.6287 & 0.6305\\
\zrow
RM-Net, CE weight 1, 15 seeds, bias & 0.6260 & 0.6277 & 0.6269 & 0.6187 & 0.6071 & 0.6185 & 0.6242\\
RM-Net, CE weight 2, 15 seeds, bias & 0.6288 & 0.6219 & 0.6253 & 0.6264 & 0.6247 & 0.6173 & 0.6230\\
\midrule
\zrow
RM-Net + BERT probe fusion & 0.6256 & 0.6258 & 0.6257 & 0.6407 & 0.6338 & 0.6279 & 0.6333\\
RM-Net + BERT probe fusion + bias & 0.6350 & 0.6308 & 0.6329 & 0.6544 & 0.6368 & 0.6279 & 0.6343\\
\zrow
RM-Net + BERT+SVD fusion & 0.6231 & 0.6224 & 0.6227 & 0.6246 & 0.6223 & 0.6312 & 0.6361\\
RM-Net + BERT+SVD fusion + bias & 0.6319 & 0.6143 & 0.6231 & 0.6169 & 0.6239 & 0.6312 & 0.6361\\
\zrow
RM-Net + LoRA fusion & 0.6351 & 0.6304 & 0.6328 & \textbf{0.6600} & \underline{0.6496} & \underline{0.6099} & \underline{0.6185}\\
\hirow
\fstar\;\textbf{RM-Net + LoRA fusion + bias (final)} & 0.6339 & \textbf{0.6394} & \textbf{0.6367} & 0.6550 & 0.6341 & \underline{0.6099} & 0.6210\\
\midrule
RM-Net (CE 1, 15 seeds) + BERT probe fusion & 0.6277 & 0.6331 & 0.6304 & 0.6439 & 0.6250 & 0.6172 & 0.6240\\
\zrow
RM-Net (CE 1, 15 seeds) + BERT probe fusion + bias & \underline{0.6360} & \underline{0.6367} & \underline{0.6363} & 0.6326 & 0.6284 & 0.6172 & 0.6242\\
RM-Net (CE 1, 15 seeds) + BERT+SVD fusion & 0.6277 & 0.6248 & 0.6262 & 0.6219 & 0.6206 & 0.6185 & 0.6242\\
\zrow
RM-Net (CE 1, 15 seeds) + BERT+SVD fusion + bias & 0.6309 & 0.6268 & 0.6288 & 0.6184 & 0.6249 & 0.6185 & 0.6242\\
RM-Net (CE 1, 15 seeds) + LoRA fusion & 0.6298 & 0.6312 & 0.6305 & 0.6377 & 0.6273 & \textbf{0.6060} & \textbf{0.6139}\\
\zrow
RM-Net (CE 1, 15 seeds) + LoRA fusion + bias & \textbf{0.6379} & 0.6341 & 0.6360 & 0.6298 & 0.6118 & \textbf{0.6060} & \textbf{0.6139}\\
\bottomrule
\end{tabularx}
\caption{All 29 evaluated candidates, grouped into single models/probes, 15-seed cores, 5-seed fusions, and 15-seed fusions. Column bests are \textbf{bold}, second-bests are \underline{underlined}, and the blue shaded \fstar{} row marks the final validation-selected configuration.}
\label{tab:candidates}
\end{table*}

%% file: appendix/app_alignment.tex
\section{Raw-Video Alignment Pipeline}
\label{app:alignment}

This appendix documents the standalone word-level alignment pipeline summarized in Section~\ref{sec:alignment-summary}. The pipeline processes 100 raw video clips together with their transcripts and operates independently of the predictive models: none of its outputs enter training, validation, or evaluation of the fused sentiment predictor.

\subsection{Common time base}
\label{app:timebase}

Let $b_a$ and $b_v$ denote the start times of the audio and video streams, with durations $T_a$ and $T_v$. The shared timeline begins at $b=\max(b_a,b_v)$ and spans $T=\min(b_a+T_a,\,b_v+T_v)-b$. Audio is decoded via FFmpeg into mono 16~kHz floating-point samples without tail padding. Video frames are decoded via PyAV, preserving original presentation timestamps (PTS) and decoded frame indices rather than deriving timestamps from a nominal frame rate. Across the 100 clips, raw video durations span 2.257 to 29.288 s, and effective common durations span 2.229 to 29.109 s, totaling 777.090 s (mean 7.771 s). Fifty clips exhibit an audio--video duration discrepancy exceeding 100 ms; every clip is restricted strictly to its common interval.

\subsection{Text features}
\label{app:text-features}

Each full transcript sentence is encoded by the frozen uncased BERT-base model (final hidden layer) without truncating to 50 positions at this stage. Words are defined as non-empty whitespace-delimited spans, excluding punctuation-only spans. For the $i$-th word covering character span $[c_i^s,c_i^e)$, let $\mathcal S_i$ be the set of overlapping WordPiece tokens. The word embedding averages their hidden states:
\begin{equation*}
\mathbf x_i^t=\frac1{|\mathcal S_i|}\sum_{u\in\mathcal S_i}\mathbf h_u\in\mathbb R^{768}.
\end{equation*}
Word-internal punctuation may contribute subword tokens to $\mathcal S_i$. Word vectors are stored in half precision, incurring a maximum absolute quantization error of 0.003899 relative to single precision.

\subsection{Acoustic features}
\label{app:acoustic}

Acoustic descriptors are extracted on a 25 ms analysis window (400 samples) with a 10 ms hop (160 samples). Given $S$ decoded samples, the count of complete frames is $N^a=\lfloor (S-400)/160\rfloor+1$, with frame $k$ centered at $\tau_k^a=(160k+200)/16000=0.0125+0.01k$ s and no trailing partial-window padding. Table~\ref{tab:acoustic} details all 74 channels. Because underlying extractors use differing native grids, openSMILE and eGeMAPS features \citep{eyben2010opensmile,eyben2016gemaps} are linearly interpolated by timestamp onto the 10 ms grid, whereas F0 and mel-cepstral channels take the nearest WORLD frame \citep{morise2016world}. Spectral and cepstral descriptors are computed with librosa \citep{mcfee2015librosa}, and mel-cepstral coefficients use SPTK via pysptk 1.0.1 \citep{sptk}. Outputs are stored in float32 with non-finite replacement counting (zero replacements occurred across all clips); semitone pitch deltas are zeroed across voiced/unvoiced transitions.

\begin{table*}[tp]
\centering
\small
\renewcommand{\arraystretch}{1.08}
\begin{tabularx}{\linewidth}{@{}lYr>{\raggedright\arraybackslash}p{2.6cm}@{}}
\toprule
\textbf{Group} & \textbf{Content} & \textbf{Dim.} & \textbf{Main tools}\\
\midrule
\textbf{Prosody} & F0, semitones re.\ 27.5\,Hz, semitone delta, voicing, RMS, RMS delta, loudness, zero-crossing rate & 8 & WORLD, waveform, openSMILE\\
\zrow
\textbf{Voice quality} & HNR, jitter, shimmer, H1--H2, alpha ratio, Hammarberg index & 6 & eGeMAPSv02\\
\textbf{Spectral shape} & centroid, bandwidth, 85\% rolloff, spectral flux, flatness, slopes (0--500, 500--1500\,Hz), entropy & 8 & librosa, openSMILE\\
\zrow
\textbf{Formants} & F1--F3 frequencies and bandwidths & 6 & eGeMAPSv02\\
\textbf{Cepstrum} & MFCC1--13 with 13 deltas (9-frame window); MCEP0--19 & 46 & librosa, WORLD, SPTK\\
\bottomrule
\end{tabularx}
\caption{The 74 acoustic feature dimensions. The complete schema is stored alongside the outputs.}
\label{tab:acoustic}
\end{table*}

\subsection{Visual features}
\label{app:visual}

Video streams are sampled onto a 25 Hz grid $\tau_j^v=j/25$ s by selecting the decoded frame whose PTS is nearest to each grid timestamp. Frames are aspect-preservingly downscaled to at most 640 pixels wide and processed by MediaPipe Face Landmarker v1 \citep{lugaresi2019mediapipe} in IMAGE mode (detection and presence thresholds 0.5; up to five faces per frame). The initial valid frame tracks the largest detected face; subsequent frames retain a face whose bounding box achieves $\text{IoU}\ge 0.5$ with the prior box, falling back to the largest face otherwise. Frames without a detected face receive a zero 35-dimensional vector and a face mask of zero. Table~\ref{tab:visual} summarizes all 35 channels. The 27 expression coefficients are descriptive outputs of the frozen landmarker, not facial action units, emotion probabilities, or affective labels.

\begin{table*}[tp]
\centering
\small
\renewcommand{\arraystretch}{1.08}
\begin{tabularx}{\linewidth}{@{}lYr@{}}
\toprule
\textbf{Group} & \textbf{Content} & \textbf{Dim.}\\
\midrule
\textbf{Expression coefficients} & brows (5), cheeks (2), eye blink, squint, and widening (6), jaw (1), smile, mouth corner, and lip pressing (6), lip pucker (1), mouth stretch (2), upper-lip raise (2), nose wrinkle (2) & 27\\
\zrow
\textbf{Head pose} & Euler angles from the facial transformation matrix (rad) & 3\\
\textbf{Geometry and dynamics} & normalized face-box area, eye aspect ratio, mouth aspect ratio, relative rotation speed between valid frames, fraction of landmarks inside the frame & 5\\
\bottomrule
\end{tabularx}
\caption{The 35 visual feature dimensions.}
\label{tab:visual}
\end{table*}

\subsection{Forced alignment and the time-quality gate}
\label{app:ctc}

Transcripts undergo normalization strictly for CTC alignment---uppercasing, retaining apostrophes, and spelling out numerals---while preserving original text and character offsets. A wav2vec~2.0 Base encoder pretrained on 960 hours of speech \citep{baevski2020wav2vec} produces frame posteriors $p_k(\cdot\mid u)$ at a 320-sample stride and 400-sample receptive field, centering CTC frame $k$ at $0.0125+0.02k$ s (at 16~kHz). Forced alignment solves
\begin{equation*}
\pi^*=\operatorname*{arg\,max}_{\pi:\,\mathcal B(\pi)=\widetilde W}\ \sum_{k}\log p_k(\pi_k\mid u),
\end{equation*}
where $\mathcal B$ collapses repeated symbols and removes blanks \citep{graves2006ctc}. Dynamic programming permits staying in state, advancing one state, or skipping a blank between distinct characters, while repeated characters must traverse a blank. Word boundaries span the first and last assigned frame centers extended by 10 ms and clipped to the valid interval. Each word's posterior $c_i$ averages its non-blank frame posteriors, and greedy decoding yields a clip-level word error rate (WER) used solely as a consistency diagnostic.

A word passes the time-quality gate only when its word posterior, clip-mean posterior, and clip WER all clear fixed thresholds set before inspecting any sentiment labels:
\begin{equation*}
\begin{aligned}
q_i&=\mathbf 1\{c_i\ge0.5\}\,\mathbf 1\{\bar c_n\ge0.5\}\\
&\quad\cdot\mathbf 1\{\mathrm{WER}_n\le0.5\},\\
\bar c_n&=\frac1{m_n}\sum_i c_i.
\end{aligned}
\end{equation*}
Words failing the gate retain their candidate spans and features, but their cross-modal use mask is set to zero. Passing this heuristic gate does not verify agreement with human boundary annotations.

\subsection{Word-level pooling and index mapping}
\label{app:pooling}

For word $i$ with candidate span $[s_i,e_i)$, let $S_i=320k_i^s+40$ and $E_i=\min\{320(k_i^e-1)+360,N_s\}$, where $k_i^s$ and $k_i^e$ are the start and exclusive end CTC frame indices and $N_s$ is the decoded sample count. The corresponding audio and video index sets are
\begin{align*}
\mathcal K_i&=\{k:\ S_i\le 160k+200<E_i\},\\
\mathcal J_i&=\{j:\ S_i\le 640j<E_i\}.
\end{align*}
If either set is empty, the frame nearest the word midpoint is substituted and logged as a fallback (ties break toward the smaller index). Audio channels are averaged over $\mathcal K_i$, while visual channels are weighted by the face-detection indicator $a_j^v$:
\begin{equation*}
\overline{\mathbf v}_i=\frac{\sum_{j\in\mathcal J_i} a_j^v\,\mathbf v_j}{\sum_{j\in\mathcal J_i} a_j^v},
\end{equation*}
setting $\overline{\mathbf v}_i=\mathbf 0$ whenever the denominator vanishes. The three-modality use mask is $M_i=(1,\,q_i a_i^a,\,q_i a_i^v)$. Every word records its character span, time interval, acoustic/visual frame indices, subword indices, and source PTS so features can be traced back to source frames. This trace serves as an index mapping rather than an exact inverse of pooling, and mean pooling represents a least-squares summary rather than signal reconstruction.

\subsection{Output interfaces}
\label{app:interfaces}

Outputs are serialized in three layers: L0 stores frame-level features on the native 10 ms acoustic and 25 Hz visual grids; L1 stores one aligned vector per word; and L2 stores fixed-length tensors on the 50-position BERT grid (Table~\ref{tab:layers}). In L2, each subword inherits its parent word's acoustic and visual vectors while \texttt{[CLS]} and \texttt{[SEP]} receive zero vectors; sequences exceeding 50 positions keep the first 49 positions plus the final \texttt{[SEP]} (affecting six clips). Non-aligned chunking aggregates fixed blocks of $r_a=\lceil 2909/500\rceil=6$ acoustic frames ($\approx$60 ms) and $r_v=\lceil 728/500\rceil=2$ visual frames ($\approx$80 ms), covering the longest clip. Metadata fields record retained BERT positions, total subwords, and total words separately.

\begin{table*}[tp]
\centering
\small
\renewcommand{\arraystretch}{1.08}
\begin{tabular}{lccc}
\toprule
\textbf{File} & \textbf{Text} & \textbf{Audio} & \textbf{Vision}\\
\midrule
\textbf{L2 aligned} (50 positions) & $(100,50,768)$ & $(100,50,74)$ & $(100,50,35)$\\
\zrow
\textbf{L2 non-aligned} (500 blocks) & $(100,50,768)$ & $(100,500,74)$ & $(100,500,35)$\\
\bottomrule
\end{tabular}
\caption{L2 output tensor shapes across the 100 raw clips.}
\label{tab:layers}
\end{table*}

\subsection{Quality checks}
\label{app:quality}

Table~\ref{tab:qc} reports quality-control diagnostics over all 100 clips. Face-detection rate is frame-weighted, and mean CTC posterior is an internal model score rather than an alignment accuracy metric. Because high WER, low posterior, missing faces, and duration mismatches can co-occur, failure counts are non-exclusive; low-scoring clips are preserved together with their diagnostic flags.

\begin{table*}[tp]
\centering
\small
\renewcommand{\arraystretch}{1.06}
\begin{tabularx}{\linewidth}{@{}Yr@{}}
\toprule
\textbf{Check or diagnostic} & \textbf{Result}\\
\midrule
Clips with all modality files and indices & \cmark\;\textbf{100 / 100}\\
\zrow
Clips passing the interface and index checks & \cmark\;\textbf{100 / 100}\\
Acoustic NaN/Inf replacements & \cmark\;0\\
\midrule
\zrow
Frame-weighted face detection rate & 81.97\%\\
Word-level visual availability before the time gate & 82.61\%\\
\zrow
Clips with face detection below 20\% (of which none detected) & 17 (13)\\
Mean CTC posterior over clips (model score) & 0.7617\\
\zrow
Clips with mean CTC posterior below 0.5 & 20\\
Median greedy WER against the transcript & 0.3417\\
\zrow
Clips with WER above 0.5 & 35\\
\midrule
\hirow
\textbf{Clips with at least one gated word} & \textbf{65 / 100}\\
\hirow
\textbf{Gated words} & \textbf{1,316 / 1,926 (68.33\%)}\\
\midrule
Clips with stream-duration difference above 100 ms & 50\\
\zrow
Fallback frames (audio / visual) & 0 / 29\\
Clips with L2 subword truncation & 6\\
\zrow
Repeated extraction: identical files, maximum absolute difference & \cmark\;\textbf{500 / 500, 0}\\
\bottomrule
\end{tabularx}
\caption{Quality checks across the 100 raw clips. Highlighted rows summarize time-quality gate coverage; \cmark{} marks exact structural/reproducibility passes.}
\label{tab:qc}
\end{table*}

\begin{figure*}[tp]
\centering
\includegraphics[width=0.95\textwidth]{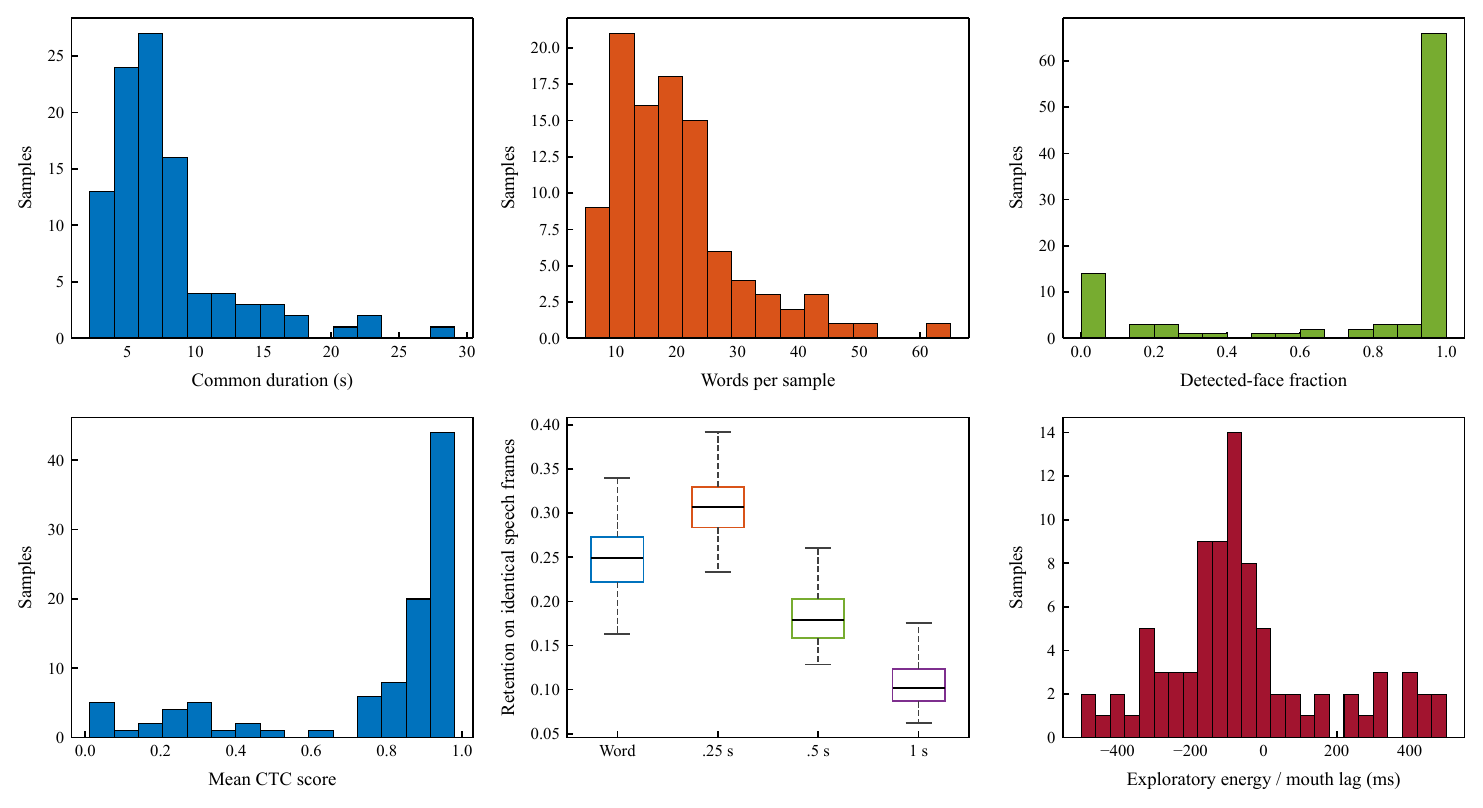}
\caption{Quality diagnostics across the 100 raw clips: common-timeline duration, words per clip, detected-face fraction, mean CTC posterior, and retained acoustic variance by grouping (65 clips with at least one gated word). Lower right: exploratory audio--visual lag histogram, not used to shift timelines.}
\label{fig:alignment-quality}
\end{figure*}

Figure~\ref{fig:alignment-quality} visualizes these diagnostic distributions. Its lower-right panel shows an exploratory audio--visual lag scan across 85 clips with sufficient paired frames, searching the cross-correlation between speech energy and mouth-opening ratio over $\pm 480$ ms. Only 15 clips peak within $\pm 40$ ms and the median absolute lag is 160 ms; given this dispersion, the pipeline applies no automatic timeline shifts.

\subsection{Cross-check with the supplied aligned split and retention}
\label{app:official}

Eighteen of the 100 raw clips also appear in the supplied aligned split (11 training, 7 held-out), exhibiting identical transcripts and exact label agreement (verified strictly for correspondence, never used in training). Table~\ref{tab:retention} compares acoustic variance retention across grouping schemes over the same frames on the 65 clips containing at least one gated word.

\begin{table}[tp]
\centering
\small
\renewcommand{\arraystretch}{1.08}
\setlength{\tabcolsep}{4pt}
\begin{tabular}{@{}lcc@{}}
\toprule
\textbf{Grouping} & \shortstack{\textbf{Retention}\\\textbf{$R$ ($\uparrow$)}} & \shortstack{\textbf{Mean non-empty}\\\textbf{groups}}\\
\midrule
\hirow
\fstar\;\textbf{Word level} & \underline{0.2489} & 20.25\\
0.25 s windows & \textbf{0.3120} & 25.75\\
\zrow
0.5 s windows & 0.1848 & 14.03\\
1.0 s windows & 0.1069 & 7.82\\
\bottomrule
\end{tabular}
\caption{Acoustic variance retained by each grouping, averaged over the 65 clips with at least one gated word. $R$ is a descriptive variance summary rather than sentiment accuracy. \fstar{} marks word-level grouping.}
\label{tab:retention}
\end{table}

\subsection{A typical clip}
\label{app:typical}

\begin{figure*}[tp]
\centering
\includegraphics[width=\textwidth]{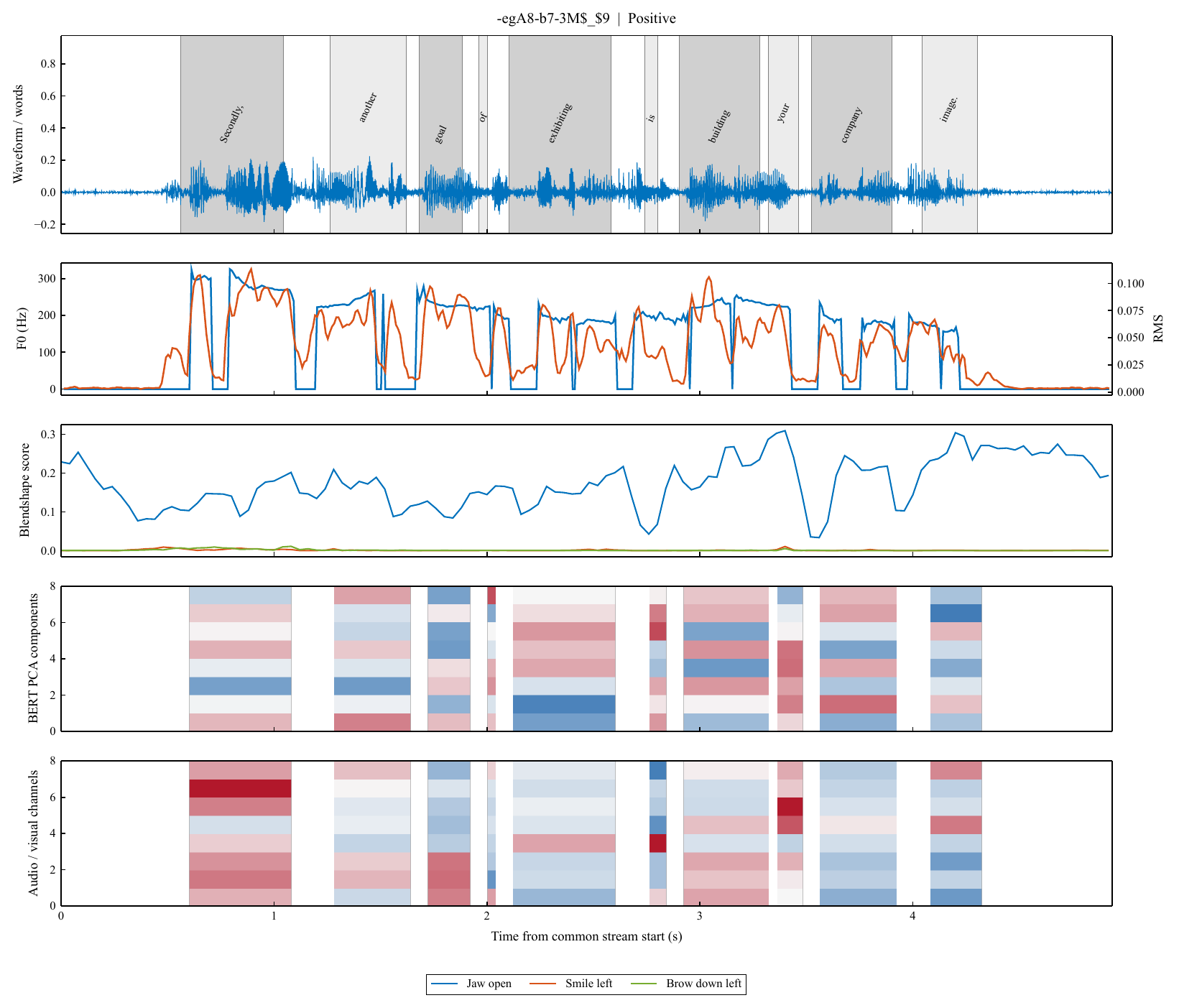}
\caption{Word-level alignment for a representative clip (\texttt{-egA8-b7-3M\_9}, Positive, 4.9\,s). Top to bottom: waveform with CTC-estimated word spans; F0 and RMS; three MediaPipe blendshape scores; principal-component projections of BERT word vectors; and per-word means of selected audio and visual channels. Shaded regions mark word boundaries, and each panel uses its own vertical scale.}
\label{fig:alignment-example}
\end{figure*}

Figure~\ref{fig:alignment-example} illustrates clip \texttt{-egA8-b7-3M\_9} (labeled Positive; common duration 4.936 s; mean CTC posterior 0.978; WER 0), in which all ten words clear the quality gate. Table~\ref{tab:typical} details its per-word mappings (0-based frame indices, half-open time intervals, and inclusive frame ranges for readability).

\begin{table*}[tp]
\centering
\small
\renewcommand{\arraystretch}{1.06}
\setlength{\tabcolsep}{5pt}
\begin{tabular}{llcccc}
\toprule
\textbf{Word} & \textbf{Time interval (s)} & \textbf{Audio frames} & \textbf{Video grid frames} & \textbf{BERT subwords} & \textbf{F0 (Hz)}\\
\midrule
Secondly, & [0.5625, 1.0425) & 55--102 & 15--26 & 1, 2 & 211.00\\
\zrow
another & [1.2625, 1.6225) & 125--160 & 32--40 & 3 & 154.51\\
goal & [1.6825, 1.8825) & 167--186 & 43--47 & 4 & 234.28\\
\zrow
of & [1.9625, 2.0025) & 195--198 & 50 & 5 & 219.63\\
exhibiting & [2.1025, 2.5825) & 209--256 & 53--64 & 6 & 130.28\\
\zrow
is & [2.7425, 2.8025) & 273--278 & 69--70 & 7 & 196.16\\
building & [2.9025, 3.2825) & 289--326 & 73--82 & 8 & 216.75\\
\zrow
your & [3.3225, 3.4625) & 331--344 & 84--86 & 9 & 176.94\\
company & [3.5225, 3.9025) & 351--388 & 89--97 & 10 & 123.54\\
\zrow
image. & [4.0425, 4.3025) & 403--428 & 102--107 & 11, 12 & 108.01\\
\bottomrule
\end{tabular}
\caption{Word-level mapping for the representative clip. Per-word jaw-opening coefficients are stored with the outputs and omitted here for brevity.}
\label{tab:typical}
\end{table*}

This example traces every word to its acoustic frames, visual grid frames, and aligned features along a shared timeline. It is illustrative rather than universal: 35 of the 100 clips yield no gated words and 13 contain no detected faces, reflecting persistent raw-media challenges.

%% file: appendix/app_explanation.tex
\section{Explanation Details}
\label{app:explanation}

This appendix expands the explanation analysis in Section~\ref{sec:results}. All attributions measure how the frozen fused predictor depends on inputs under the specified interventions; they do not establish human emotional causation and have not been evaluated against human evidence annotations.

\subsection{Targets and attributions}
\label{app:targets}

Each utterance is audited under two targets: the fused intensity $\hat y$ and the probability of the originally predicted class $p_{\hat c_0}$, where $\hat c_0$ is fixed on the unperturbed input and held constant under perturbation even if the argmax class flips. Treating the three modalities as players in a Shapley game, the value $F(S)$ of coalition $S\subseteq\{t,a,v\}$ is the complete fused predictor's output when only the observations of $S$ are retained. Modality $m$'s exact Shapley attribution is
\begin{equation*}
\begin{aligned}
\phi_m&=\sum_{S\subseteq\{t,a,v\}\setminus\{m\}}
\frac{|S|!\,(3-|S|-1)!}{3!}\\
&\qquad\cdot\bigl(F(S\cup\{m\})-F(S)\bigr).
\end{aligned}
\end{equation*}
A modality's relative share is $|\phi_m|/\sum_{m'}|\phi_{m'}|$, with the largest share designated dominant (ties are flagged). Empty coalitions evaluate to the model's learned empty-set prior rather than zero, ensuring the efficiency identity holds against the same baseline.

Position-level evidence is obtained via single-position occlusion. Each observed content position is occluded individually under four intervention scopes (text, audio, vision, and all three modalities jointly at that position), recomputing the complete fused output each time. Positions are ranked by $|\Delta y|$ for intensity and by the decrease in $p_{\hat c_0}$ for classification, breaking ties toward the smaller position index. Special tokens are excluded from ranking while remaining intact in the predictor.

\subsection{Fidelity protocol and flips}
\label{app:fidelity}

Deletion and retention curves remove or keep the top fraction $\rho\in\{0,0.1,\dots,0.8\}$ of content positions ($\lceil\rho K\rceil$ out of $K$ content positions) and recompute the full fused predictor. Random baselines evaluate five random permutations independently before averaging metrics. Curves are integrated over $[0,0.8]$ via the trapezoidal rule and normalized by 0.8. Intensity fidelity measures absolute change $|\Delta y|$; classification fidelity measures the signed decrease in $p_{\hat c_0}$, so a negative retention AUC indicates that keeping top-ranked positions increases predicted-class probability on average.

Class-flip analysis removes content positions sequentially along the classification ranking and records the first deletion count that alters the three-class prediction, censoring non-flipping utterances at $K+1$. This captures the first flip along a fixed ranking rather than the global minimum over arbitrary subsets, applying the identical censoring rule to random permutations. Across the 728 validation utterances, the median censored first-flip count is 11 for the selected ranking versus 18 for random permutations (averaging the five permutations within each utterance before taking the cross-utterance median), with 37.36\% of utterances never flipping under the selected order.

\subsection{Numerical checks}
\label{app:numerical}

Exact Shapley attributions satisfy the efficiency identity within maximum residuals of $4.44\times10^{-16}$ for intensity and $1.67\times10^{-16}$ for class probability, and decomposing fused intensity into RM-Net and text-expert terms yields a maximum residual of $1.25\times10^{-7}$. Recomputing 12 intervention outputs from raw inputs---three utterances from unlabeled set B (IDs 01, 10, and 20), each under full input, all-text-missing, local mask, and all-modalities-missing conditions---matches cached results within $2.38\times10^{-7}$ for intensity and $2.73\times10^{-8}$ for probability, with 100\% class agreement. Stored float32 text contexts were reused for speed while pooling, branch outputs, and fusion weights were recomputed on every intervention.

\subsection{Conflicts, margins, and error rates}
\label{app:conflict}

A text--vision conflict occurs when the two modalities carry opposite-sign Shapley attributions for class probability. On validation, the 357 conflicting utterances exhibit an error rate of 38.94\%, compared with 32.35\% on the remaining 371 utterances---a practical signal for manual triage, though not causal evidence. Stratifying by fused probability margin (gap between the top two probabilities) produces error rates of 48.90\% and 50.55\% in the bottom two quartiles versus 14.84\% in the top quartile, showing a strong but non-monotone trend. The 7 validation utterances where audio dominates classification have a 57.14\% error rate, too small a count to support modality-quality comparisons.

\subsection{Word deletion against encoded masking}
\label{app:real-deletion}

For each validation utterance, we removed the surface word exhibiting the largest classification-occlusion impact, re-encoded the shortened text with frozen BERT and each LoRA expert, and mapped surviving positions back to their original indices while keeping audio and vision aligned with their tokens. Positions newly exposed at the truncated tail lacked supplied audio or vision evidence and were marked unobserved (affecting 35 of 728 comparisons). Table~\ref{tab:real-deletion} compares this intervention against joint three-modality encoded masking of the same word.

\begin{table}[t]
\centering
\small
\renewcommand{\arraystretch}{1.08}
\setlength{\tabcolsep}{3pt}
\begin{tabular}{@{}lccc@{}}
\toprule
\textbf{Target} & \shortstack{\textbf{Signed}\\\textbf{$\rho$ ($\uparrow$)}} & \shortstack{\textbf{Abs.}\\\textbf{$\rho$ ($\uparrow$)}} & \shortstack{\textbf{Mean abs.}\\\textbf{diff.\ ($\downarrow$)}}\\
\midrule
Fused intensity & \cbest\textbf{0.6680} & \cbest\textbf{0.4748} & 0.1707\\
\zrow
Class prob. & \cneg\tneg{0.2903} & \cneg\tneg{0.3522} & 0.0785\\
\bottomrule
\end{tabular}
\caption{Agreement between encoded masking and surface word deletion for the top-ranked word of each validation utterance (728 paired comparisons; $\rho$: Spearman correlation). Shading contrasts moderate intensity agreement (blue) with weak probability agreement (red).}
\label{tab:real-deletion}
\end{table}

Because encoded masking preserves contextual traces of the masked word whereas real deletion also shifts token positions and alters the visible-text denominator, the comparison conflates contextual leakage, index shifts, and reliability changes. Table~\ref{tab:real-deletion} shows moderate agreement for intensity and weak agreement for class probability, underscoring that post-encoding masking must not be equated with true word absence.

\subsection{Representative misclassifications}
\label{app:cases}

Table~\ref{tab:cases} highlights three misclassified validation utterances chosen solely to illustrate distinct failure modes without affecting reported metrics.

\begin{table*}[t]
\centering
\small
\renewcommand{\arraystretch}{1.08}
\begin{tabular}{lcccc}
\toprule
\textbf{Case} & \shortstack{\textbf{True polarity /}\\\textbf{intensity}} & \shortstack{\textbf{Predicted}\\\textbf{polarity}} & \shortstack{\textbf{Predicted}\\\textbf{intensity}} & \textbf{Margin}\\
\midrule
\textbf{E1} (\texttt{f\_ZJ7L14oYQ\$\_\$21}) & \tpos{Pos} / $2.000$ & \tneg{Neg} & $-1.681$ & \textbf{0.940}\\
\zrow
\textbf{E2} (\texttt{EQxkXiWAm08\$\_\$8}) & \tpos{Pos} / $0.333$ & Neu & $0.100$ & 0.642\\
\textbf{E3} (\texttt{x2lOwQaAn4g\$\_\$4}) & Neu / $0.000$ & \tpos{Pos} & $1.675$ & 0.863\\
\bottomrule
\end{tabular}
\caption{Three illustrative misclassified validation utterances. Margin denotes the gap between the top two fused class probabilities.}
\label{tab:cases}
\end{table*}

In E1, the speaker praises a product as more convenient than boots that take forever to put on; negative phrasing targeting the compared boots appears to mislead the model into predicting strong negative sentiment. Its top three class-supporting positions correspond to ``aren't'' and two instances of ``just'', each causing a small probability drop and pointing toward comparative phrasing as a hypothesis for review rather than an established cause. E2 is a mildly positive factual remark classified as Neutral while receiving a weakly positive intensity (0.100), illustrating boundary ambiguity between weak Positive and Neutral. E3 is labeled Neutral yet describes convenience favorably; its top-ranked word, ``guarantee'', carries a negative probability effect (masking it slightly raises the predicted-class probability), showing that single-position rankings must be interpreted with both sign and context.

\subsection{Set B, sample 19}
\label{app:sample19}

Sample 19 of unlabeled set B serves as our reference dual-target explanation case. It is predicted Negative with fused intensity 0.034 and probability margin 0.270, illustrating cross-target divergence: text dominates classification while vision dominates intensity. Classification Shapley values are $0.25394$ (text), $-0.05226$ (audio), and $-0.02965$ (vision), showing lexical support for Negative partially offset by speech and vision. Intensity Shapley values are $-0.25695$ (text), $0.01894$ (audio), and $0.26464$ (vision), so text and vision nearly cancel around zero. Because Shapley values are anchored to the empty-set baseline, dominant-modality labels alone do not specify prediction polarity.

\begin{figure*}[t]
\centering
\includegraphics[width=0.92\textwidth]{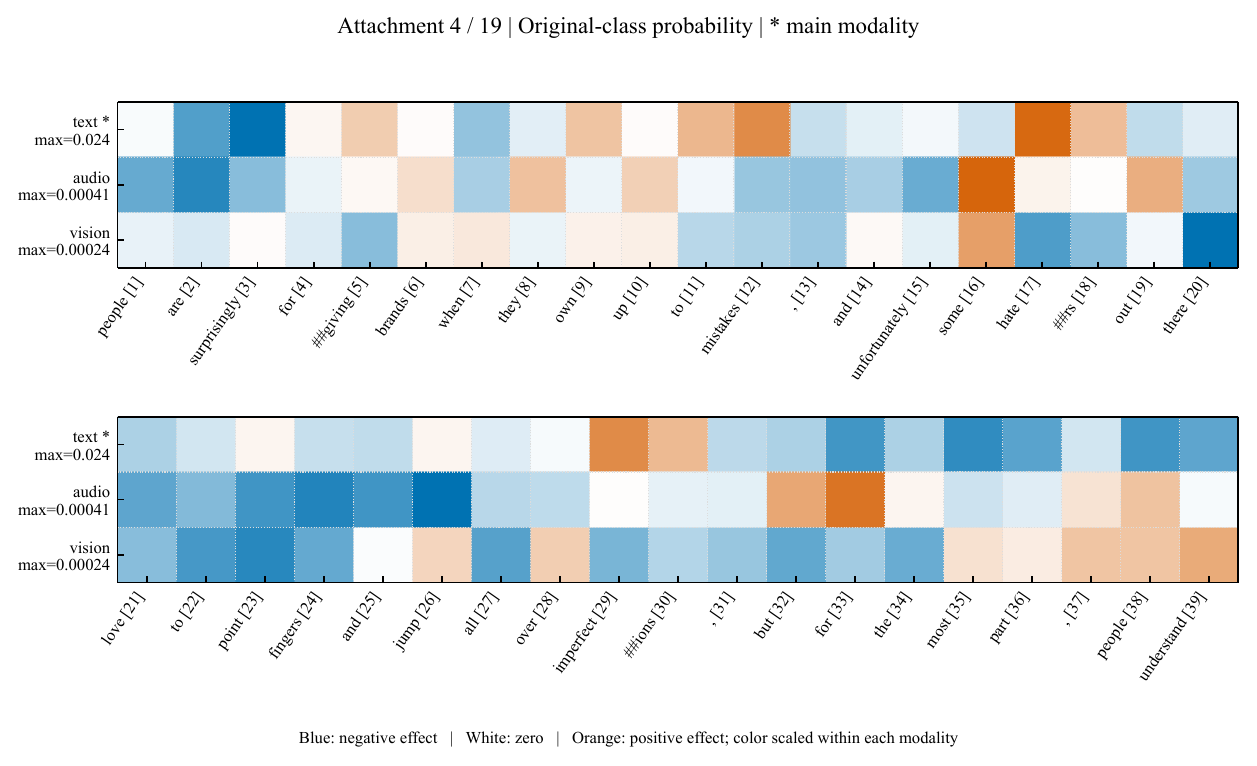}
\caption{Set B, sample 19: single-position occlusion effects on the classification target (probability of the originally predicted class). The figure title preserves the release identifier ``Attachment 4 / 19''.}
\label{fig:sample19-cls}
\end{figure*}

\begin{figure*}[t]
\centering
\includegraphics[width=0.92\textwidth]{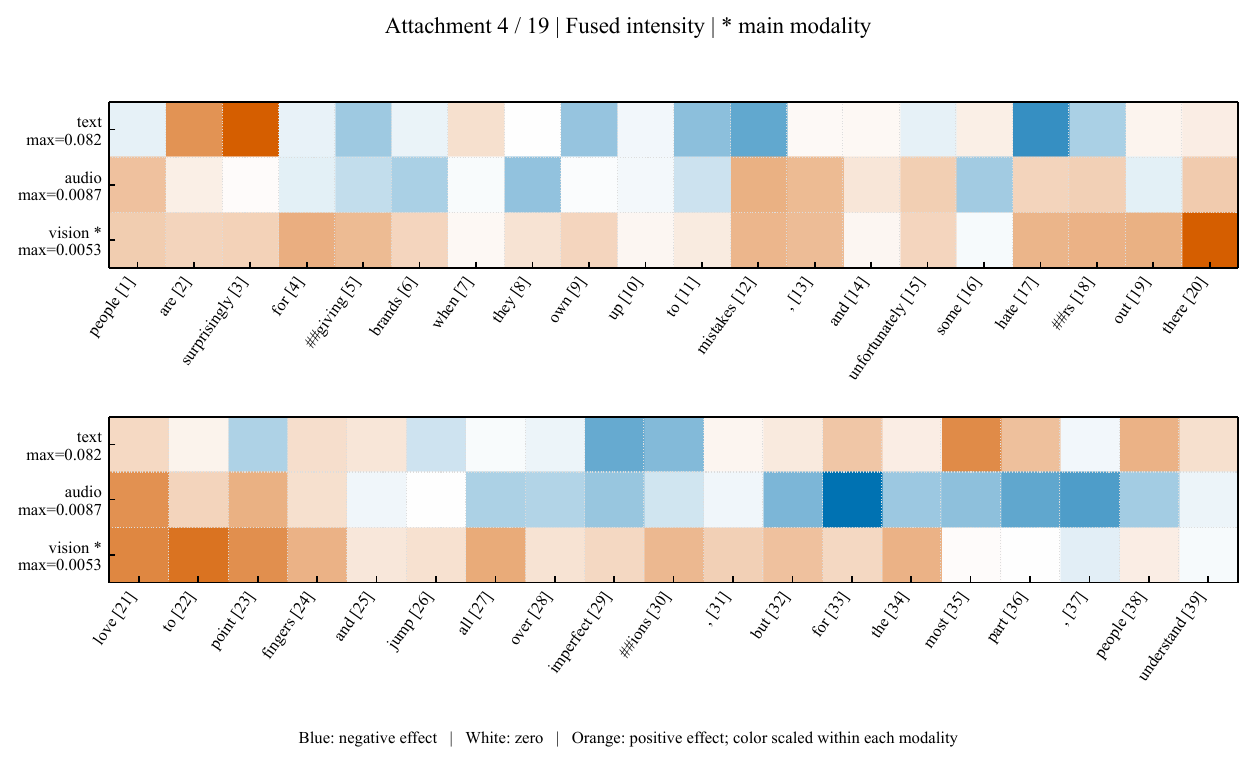}
\caption{Set B, sample 19: single-position occlusion effects on fused intensity, where vision is the dominant modality. Colors are scaled within each modality as in Figure~\ref{fig:sample19-cls}.}
\label{fig:sample19-int}
\end{figure*}

Figures~\ref{fig:sample19-cls} and~\ref{fig:sample19-int} display single-position occlusion effects across all 39 valid content positions for both targets. Colors are scaled independently within each modality row---comparing positions within a modality rather than absolute magnitudes across modalities---with blue indicating negative effects, orange positive effects, and an asterisk marking the dominant modality.

\subsection{Scope of the explanations}
\label{app:scope}

These explanations are exact for both the three-player Shapley decomposition and the specified occlusion interventions. Cross-seed stability and parameter-randomization checks for the fused predictor remain future work. Furthermore, the attributions do not identify human-annotated rationales, and any word-level timestamps inherit the alignment boundaries documented in Appendix~\ref{app:alignment}.

%% file: appendix/app_predictions.tex
\section{Predictions on the Unlabeled Sets}
\label{app:predictions}

Two unlabeled sets are used strictly for inference and descriptive analysis: set A contains 30 utterances with absent or zero-valued spans in the aligned release (release attachment~3), and set B contains 20 video-backed utterances used for dual-target explanations (release attachment~4). Because neither set includes ground-truth labels, no accuracy metrics are reported; release names are retained solely for traceability.

\subsection{Set A}
\label{app:set-a}

\begin{figure*}[tp]
\centering
\includegraphics[width=0.92\textwidth]{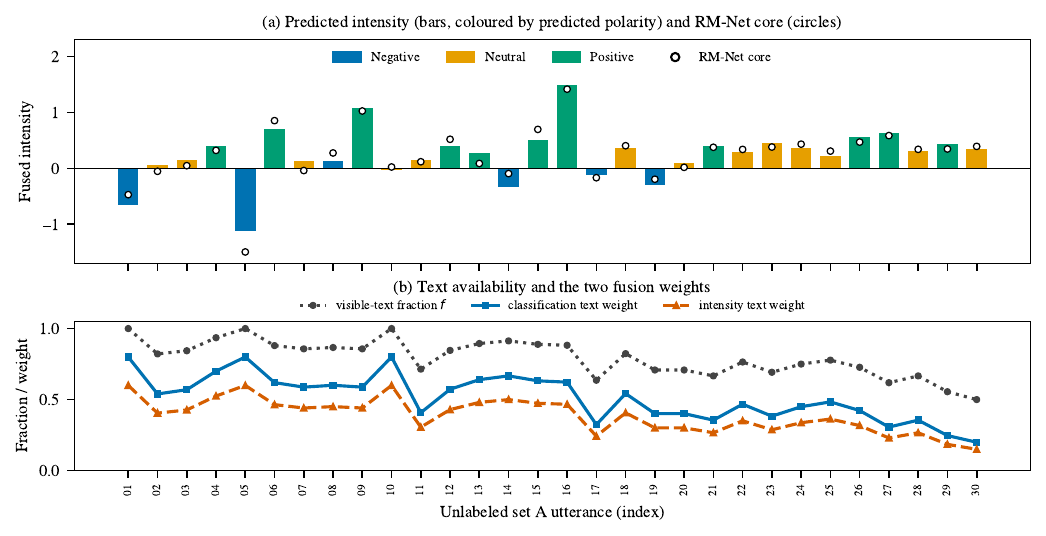}
\caption{Unlabeled set A diagnostics. (a) Fused intensity bars colored by predicted polarity, overlaid with RM-Net core intensity (open circles). (b) Visible-text fraction $f$ and corresponding classification and intensity text-expert weights per utterance.}
\label{fig:unlabeled-a}
\end{figure*}

Table~\ref{tab:unlabeled-a} lists the fused predictions for set A, and Figure~\ref{fig:unlabeled-a} plots them alongside the visible-text fraction $f$ and both fusion weights. The fused model predicts 6 Negative, 13 Neutral, and 11 Positive utterances, with $f$ ranging from 0.50 to 1.00 and 27 utterances containing \texttt{[UNK]} tokens (Appendix~\ref{app:availability}). Each row reports fused class probabilities together with RM-Net core intensity for direct comparison. Because expert weights follow $w=0.8f^2$ and $v=0.6f^2$, they vanish only when no content text is visible.

\input{appendix/tab_unlabeled_a}

\subsection{Set B}
\label{app:set-b}

Table~\ref{tab:unlabeled-b} summarizes predictions and dominant modalities for set B: 8 Negative, 5 Neutral, and 7 Positive utterances. Text dominates classification in all 20 utterances, whereas intensity is dominated by text in 16 cases and by vision in 4. Table~\ref{tab:evidence-b} lists the highest-ranked evidence position per modality together with its estimated time interval when available. Timing for samples 15 and 16 failed the time-quality gate, and samples 01 and 13 contain no detected faces; these flags are recorded explicitly in the table and should be retained whenever inspecting these explanations.

\input{appendix/tab_unlabeled_b}

\input{appendix/tab_evidence_b}

%% file: appendix/tab_unlabeled_a.tex
\begin{table*}[tp]
\centering
\small
\renewcommand{\arraystretch}{1.04}
\setlength{\tabcolsep}{4pt}
\begin{tabular}{lcrrcccrr}
\toprule
& & \multicolumn{2}{c}{\textbf{Intensity}} & \multicolumn{3}{c}{\textbf{Fused probabilities}} & \multicolumn{2}{c}{\textbf{Weights}}\\
\cmidrule(lr){3-4}\cmidrule(lr){5-7}\cmidrule(lr){8-9}
\textbf{Set A} & \textbf{Polarity} & Fused & RM core & $p_{\mathrm{neg}}$ & $p_{\mathrm{neu}}$ & $p_{\mathrm{pos}}$ & $w_{\mathrm{cls}}$ & $w_{\mathrm{reg}}$\\
\midrule
A01 & \tneg{Negative} & -0.660 & -0.470 & \textbf{0.697} & 0.185 & 0.118 & 0.80 & 0.60\\
\zrow
A02 & Neutral & +0.061 & -0.052 & 0.199 & \textbf{0.647} & 0.154 & 0.54 & 0.40\\
A03 & Neutral & +0.152 & +0.048 & 0.187 & \textbf{0.558} & 0.255 & 0.57 & 0.43\\
\zrow
A04 & \tpos{Positive} & +0.398 & +0.322 & 0.093 & 0.333 & \textbf{0.573} & 0.70 & 0.53\\
A05 & \tneg{Negative} & -1.127 & -1.494 & \textbf{0.849} & 0.091 & 0.061 & 0.80 & 0.60\\
\zrow
A06 & \tpos{Positive} & +0.710 & +0.853 & 0.047 & 0.364 & \textbf{0.589} & 0.62 & 0.46\\
A07 & Neutral & +0.137 & -0.039 & 0.211 & \textbf{0.411} & 0.378 & 0.59 & 0.44\\
\zrow
A08 & \tneg{Negative} & +0.138 & +0.276 & \textbf{0.479} & 0.304 & 0.218 & 0.60 & 0.45\\
A09 & \tpos{Positive} & +1.087 & +1.026 & 0.011 & 0.116 & \textbf{0.873} & 0.59 & 0.44\\
\zrow
A10 & Neutral & -0.035 & +0.025 & 0.149 & \textbf{0.764} & 0.086 & 0.80 & 0.60\\
A11 & Neutral & +0.154 & +0.119 & 0.158 & \textbf{0.495} & 0.346 & 0.41 & 0.31\\
\zrow
A12 & \tpos{Positive} & +0.399 & +0.520 & 0.250 & 0.329 & \textbf{0.421} & 0.57 & 0.43\\
A13 & \tpos{Positive} & +0.270 & +0.088 & 0.164 & 0.403 & \textbf{0.433} & 0.64 & 0.48\\
\zrow
A14 & \tneg{Negative} & -0.329 & -0.093 & \textbf{0.701} & 0.232 & 0.067 & 0.67 & 0.50\\
A15 & \tpos{Positive} & +0.505 & +0.697 & 0.036 & 0.360 & \textbf{0.605} & 0.63 & 0.47\\
\zrow
A16 & \tpos{Positive} & +1.490 & +1.415 & 0.039 & 0.068 & \textbf{0.893} & 0.62 & 0.47\\
A17 & \tneg{Negative} & -0.110 & -0.167 & \textbf{0.450} & 0.408 & 0.142 & 0.32 & 0.24\\
\zrow
A18 & Neutral & +0.357 & +0.405 & 0.120 & \textbf{0.485} & 0.395 & 0.54 & 0.41\\
A19 & \tneg{Negative} & -0.295 & -0.194 & \textbf{0.492} & 0.395 & 0.113 & 0.40 & 0.30\\
\zrow
A20 & Neutral & +0.102 & +0.018 & 0.217 & \textbf{0.578} & 0.205 & 0.40 & 0.30\\
A21 & \tpos{Positive} & +0.404 & +0.377 & 0.046 & 0.356 & \textbf{0.597} & 0.36 & 0.27\\
\zrow
A22 & Neutral & +0.302 & +0.339 & 0.214 & \textbf{0.397} & 0.389 & 0.47 & 0.35\\
A23 & Neutral & +0.447 & +0.381 & 0.054 & \textbf{0.559} & 0.387 & 0.38 & 0.29\\
\zrow
A24 & Neutral & +0.359 & +0.433 & 0.336 & \textbf{0.345} & 0.319 & 0.45 & 0.34\\
A25 & Neutral & +0.219 & +0.309 & 0.119 & \textbf{0.538} & 0.342 & 0.48 & 0.36\\
\zrow
A26 & \tpos{Positive} & +0.559 & +0.470 & 0.079 & 0.263 & \textbf{0.658} & 0.42 & 0.32\\
A27 & \tpos{Positive} & +0.633 & +0.586 & 0.069 & 0.284 & \textbf{0.647} & 0.31 & 0.23\\
\zrow
A28 & Neutral & +0.316 & +0.340 & 0.052 & \textbf{0.532} & 0.416 & 0.36 & 0.27\\
A29 & \tpos{Positive} & +0.434 & +0.347 & 0.117 & 0.401 & \textbf{0.482} & 0.25 & 0.19\\
\zrow
A30 & Neutral & +0.349 & +0.393 & 0.202 & \textbf{0.557} & 0.242 & 0.20 & 0.15\\
\bottomrule
\end{tabular}
\caption{Predictions for unlabeled set A (30 utterances). Polarity and fused intensity are the fused outputs; RM core is core intensity; the highest fused class probability per row is \textbf{bold}; $w_{\mathrm{cls}}$ and $w_{\mathrm{reg}}$ are the classification and intensity text-expert weights. No labels are available.}
\label{tab:unlabeled-a}
\end{table*}

%% file: appendix/tab_unlabeled_b.tex
\begin{table*}[tp]
\centering
\small
\renewcommand{\arraystretch}{1.05}
\setlength{\tabcolsep}{4pt}
\begin{tabular}{@{}lcrcccc@{}}
\toprule
& & & \multicolumn{2}{c}{\textbf{Main modality}} & \multicolumn{2}{c}{\textbf{Shapley shares T/A/V (\%)}}\\
\cmidrule(lr){4-5}\cmidrule(lr){6-7}
\textbf{Set B} & \textbf{Polarity} & \textbf{Intensity} & Class & Intensity & Class & Intensity\\
\midrule
B01 & Neutral & +0.369 & text & text & 65.4 / 19.4 / 15.2 & 89.6 / 6.2 / 4.2\\
\zrow
B02 & \tpos{Positive} & +0.250 & text & text & 66.3 / 20.8 / 12.9 & 70.6 / 18.1 / 11.3\\
B03 & \tneg{Negative} & -0.350 & text & text & 78.7 / 8.9 / 12.5 & 80.1 / 18.6 / 1.3\\
\zrow
B04 & \tneg{Negative} & -0.358 & text & text & 75.5 / 9.9 / 14.7 & 58.6 / 20.7 / 20.7\\
B05 & Neutral & +0.401 & text & \textbf{\tbest{vision}} & 66.9 / 12.3 / 20.8 & 19.0 / 19.1 / \textbf{61.9}\\
\zrow
B06 & \tpos{Positive} & +0.755 & text & text & 79.0 / 1.5 / 19.5 & 68.5 / 13.5 / 17.9\\
B07 & \tpos{Positive} & +0.769 & text & text & 83.6 / 7.3 / 9.1 & 64.7 / 12.5 / 22.8\\
\zrow
B08 & \tpos{Positive} & +0.585 & text & text & 80.2 / 6.1 / 13.7 & 74.9 / 10.5 / 14.6\\
B09 & \tneg{Negative} & -2.275 & text & text & 94.2 / 5.0 / 0.8 & 98.0 / 0.9 / 1.1\\
\zrow
B10 & \tneg{Negative} & -1.705 & text & text & 87.9 / 5.4 / 6.6 & 92.5 / 4.5 / 3.0\\
B11 & \tneg{Negative} & -1.257 & text & text & 91.3 / 2.1 / 6.7 & 86.7 / 8.9 / 4.4\\
\zrow
B12 & \tneg{Negative} & -0.823 & text & text & 92.6 / 3.4 / 4.0 & 84.9 / 1.0 / 14.1\\
B13 & Neutral & +0.309 & text & text & 87.2 / 12.8 / 0.0 & 86.5 / 13.5 / 0.0\\
\zrow
B14 & Neutral & +0.270 & text & \textbf{\tbest{vision}} & 72.6 / 7.7 / 19.6 & 30.9 / 22.9 / \textbf{46.2}\\
B15 & \tpos{Positive} & +0.630 & text & text & 69.4 / 20.5 / 10.0 & 55.3 / 31.0 / 13.8\\
\zrow
B16 & \tneg{Negative} & -2.367 & text & text & 95.2 / 2.9 / 1.9 & 98.4 / 0.3 / 1.3\\
B17 & \tpos{Positive} & +1.579 & text & text & 83.7 / 8.0 / 8.4 & 94.0 / 1.3 / 4.6\\
\zrow
B18 & Neutral & -0.324 & text & \textbf{\tbest{vision}} & 54.3 / 13.1 / 32.6 & 15.1 / 11.6 / \textbf{73.3}\\
\hirow
B19 & \tneg{Negative} & +0.034 & text & \textbf{\tbest{vision}} & 75.6 / 15.6 / 8.8 & 47.5 / 3.5 / \textbf{49.0}\\
\zrow
B20 & \tpos{Positive} & +0.700 & text & text & 60.0 / 10.7 / 29.3 & 70.2 / 0.5 / 29.4\\
\bottomrule
\end{tabular}
\caption{Predictions and dominant modalities for unlabeled set B (20 utterances). Shares are normalized absolute Shapley values per target. Navy \textbf{bold} marks the four vision-dominated intensity cases; blue shading highlights reference case B19 (Appendix~\ref{app:sample19}).}
\label{tab:unlabeled-b}
\end{table*}

%% file: appendix/tab_evidence_b.tex
\begin{table*}[tp]
\centering
\small
\renewcommand{\arraystretch}{1.05}
\setlength{\tabcolsep}{3.5pt}
\begin{tabularx}{\linewidth}{@{}lccYYY@{}}
\toprule
\textbf{Set B} & \textbf{Polarity} & \shortstack{\textbf{Main}\\\textbf{(class)}} & \textbf{Top text evidence} & \textbf{Top audio evidence} & \textbf{Top vision evidence}\\
\midrule
B01 & Neutral & text & \textbf{when} (1.98--2.10\,s; est.) & \textbf{its} (7.04--7.14\,s; est.) & \textbf{ensuring} (4.64--5.02\,s; est.)\\
\zrow
B02 & \tpos{Positive} & text & \textbf{happiness} (1.44--1.80\,s; est.) & \textbf{want} (0.60--0.70\,s; est.) & \textbf{happiness} (1.44--1.80\,s; est.)\\
B03 & \tneg{Negative} & text & \textbf{just} (2.72--2.88\,s; est.) & \textbf{account,} (6.22--6.52\,s; est.) & \textbf{take} (7.04--7.22\,s; est.)\\
\zrow
B04 & \tneg{Negative} & text & \textbf{not.} (6.02--6.24\,s; est.) & \textbf{team} (1.28--1.44\,s; est.) & \textbf{chances} (2.90--3.24\,s; est.)\\
B05 & Neutral & text & \textbf{for} (4.16--4.26\,s; est.) & \textbf{for} (4.16--4.26\,s; est.) & \textbf{Wagon} (4.64--4.92\,s; est.)\\
\zrow
B06 & \tpos{Positive} & text & \textbf{impressive} (3.96--4.40\,s; est.) & \textbf{dance} (4.46--4.70\,s; est.) & \textbf{you're} (0.62--0.76\,s; est.)\\
B07 & \tpos{Positive} & text & \textbf{``enthusiastically} (6.00--7.02\,s; est.) & \textbf{discussed} (7.10--7.56\,s; est.) & \textbf{partnerships,} (12.90--13.48\,s; est.)\\
\zrow
B08 & \tpos{Positive} & text & \textbf{Universal} (2.04--2.50\,s; est.) & \textbf{That} (0.64--0.78\,s; est.) & \textbf{Grand} (3.38--3.68\,s; est.)\\
B09 & \tneg{Negative} & text & \textbf{not} (1.36--1.46\,s; est.) & \textbf{not} (1.36--1.46\,s; est.) & \textbf{not} (3.04--3.14\,s; est.)\\
\zrow
B10 & \tneg{Negative} & text & \textbf{terrible} (8.68--9.20\,s; est.) & \textbf{flicks} (3.90--4.18\,s; est.) & \textbf{this} (6.94--7.10\,s; est.)\\
B11 & \tneg{Negative} & text & \textbf{be} (0.78--0.86\,s; est.) & \textbf{made} (3.52--3.68\,s; est.) & \textbf{have} (3.34--3.46\,s; est.)\\
\zrow
B12 & \tneg{Negative} & text & \textbf{low.} (12.02--12.26\,s; est.) & \textbf{had} (2.50--2.70\,s; est.) & \textbf{sag,} (9.94--10.30\,s; est.)\\
B13 & Neutral & text & \textbf{more.} (8.44--8.60\,s; est.) & \textbf{the} (6.86--6.92\,s; est.) & -- (\tneg{no observed evidence})\\
\zrow
B14 & Neutral & text & \textbf{He} (0.50--0.56\,s; est.) & \textbf{(UFE)} (6.64--6.84\,s; \tneg{low trust}) & \textbf{(UFE)} (6.64--6.84\,s; \tneg{low trust})\\
B15 & \tpos{Positive} & text & \textbf{transform} (6.74--6.98\,s; \tneg{low trust}) & \textbf{However,} (0.08--0.58\,s; \tneg{low trust}) & \textbf{lives} (7.02--7.28\,s; \tneg{low trust})\\
\zrow
B16 & \tneg{Negative} & text & \textbf{terrible} (1.82--2.16\,s; \tneg{low trust}) & \textbf{It's} (0.48--0.62\,s; \tneg{low trust}) & \textbf{movie} (2.20--2.50\,s; \tneg{low trust})\\
B17 & \tpos{Positive} & text & \textbf{will} (6.80--6.94\,s; est.) & \textbf{guru.} (8.74--9.02\,s; est.) & \textbf{design} (1.76--2.12\,s; est.)\\
\zrow
B18 & Neutral & text & \textbf{first} (1.72--1.94\,s; est.) & \textbf{Denmark} (0.82--1.20\,s; est.) & \textbf{mackerel} (14.80--15.20\,s; est.)\\
\hirow
\fstar\;\textbf{B19} & \tneg{Negative} & text & \textbf{haters} (6.38--6.74\,s; est.) & \textbf{some} (6.14--6.30\,s; est.) & \textbf{some} (6.14--6.30\,s; est.)\\
\zrow
B20 & \tpos{Positive} & text & \textbf{course,} (0.82--1.04\,s; est.) & \textbf{today.} (9.46--9.68\,s; est.) & \textbf{and} (4.56--4.64\,s; est.)\\
\bottomrule
\end{tabularx}
\caption{Top-ranked evidence for unlabeled set B: highest-ranked position per modality with its time interval (seconds from common stream start) and timing status (est.: estimated gated timing; \tneg{low trust}: candidate timing failing the time-quality gate). Blue shading (\fstar) marks reference case B19.}
\label{tab:evidence-b}
\end{table*}

%% file: appendix/app_ai.tex
\section{Use of AI Assistants}
\label{app:ai}

Generative AI tools assisted with brainstorming, drafting, language polishing, coding, debugging, and typesetting. The authors designed all methods, conducted and verified all experiments, and assume full responsibility for the manuscript and released code.